\documentclass[11pt]{article}

\usepackage{acl}
\usepackage{times}
\usepackage{latexsym}
\usepackage[T1]{fontenc}
\usepackage[utf8]{inputenc}
\usepackage{microtype}
\usepackage{inconsolata}

\usepackage{graphicx}
\usepackage{subcaption}
\usepackage{amsmath}
\usepackage{amssymb}
\usepackage{mathtools}
\usepackage{amsthm}
\usepackage{booktabs}
\usepackage{threeparttable}
\usepackage{enumitem}
\usepackage{multirow}
\usepackage{colortbl}
\usepackage{wrapfig}
\usepackage{xcolor}
\usepackage{makecell}
\usepackage[capitalize,noabbrev]{cleveref}

\theoremstyle{plain}

\theoremstyle{definition}

\theoremstyle{remark}

\title{Bridging Modality Disconnect in Self-Reflection via Closed-Loop Visually Grounded Verification}

\author{
\textbf{Haoyu Zhang}$^{1,2}$ \quad
\textbf{Yuwei Wu}$^{1,2}$ \quad
\textbf{Pengxiang Li}$^{1}$ \quad
\textbf{Xintong Zhang}$^{1}$ \\
\textbf{Zhi Gao}$^{1,2,\dagger}$ \quad
\textbf{Rui Gao}$^{1,2}$ \quad
\textbf{Mingyang Gao}$^{1}$ \quad
\textbf{Che Sun}$^{2}$ \quad
\textbf{Yunde Jia}$^{2}$ \\
\\
$^{1}$Beijing Institute of Technology \\
$^{2}$Shenzhen MSU-BIT University \\
$^\dagger$Corresponding author.
}

\begin{document}

\maketitle

\begin{abstract}
Self-reflection has become a key mechanism for improving reasoning in Vision-Language Models (VLMs), yet this corrective mechanism often fails when resolving complex fine-grained visual ambiguities. This performance degradation stems from the issue of modality disconnect in self-reflection: most existing models execute self-reflection either within textual or latent space, lacking an explicit mechanism to align textual reasoning with visual evidence.
In this paper, we propose MIRROR, a closed-loop visual reflection framework comprising four steps: response generation, error identification, region-based visual verification, and revision. In this cycle, the model first generates an initial response, identifies uncertain logical statements in the response that require visual verification, then grounds them in relevant image regions, and finally revises based on the visual evidence. We construct a multi-turn visual reflection dataset ReflectV, which empowers the model with such a reflective capability. 
Extensive experiments across diverse multimodal benchmarks show that MIRROR improves performance and reduces visual hallucinations by an average of about $7.2\%$, demonstrating the advantage of transforming self-reflection from open-loop textual revision into closed-loop, visually grounded verification. The project page is at \href{https://anonymous.4open.science/w/MIRROR-0524/}{here}.
\end{abstract}

\section{Introduction}

Reasoning in Vision-Language Models (VLMs) has advanced rapidly, evolving from standard instruction following to the incorporation of Chain-of-Thought (CoT) mechanisms~\cite{cot,llava-cot} that decompose complex problems into intermediate steps. Concurrently, a series of works introduces self-reflection mechanisms that enable models to critique and revise their outputs through additional reasoning steps~\cite{cft,srpo,fire}. These methods generally follow a "critique-then-refine" protocol, either expanding textual reasoning chains or adding special reflection tokens. Nevertheless, these methods operate entirely in the textual space---the model revises its answer without re-examining the underlying visual evidence.

\begin{figure}[t]
    \centering
    \includegraphics[width=0.5\textwidth]{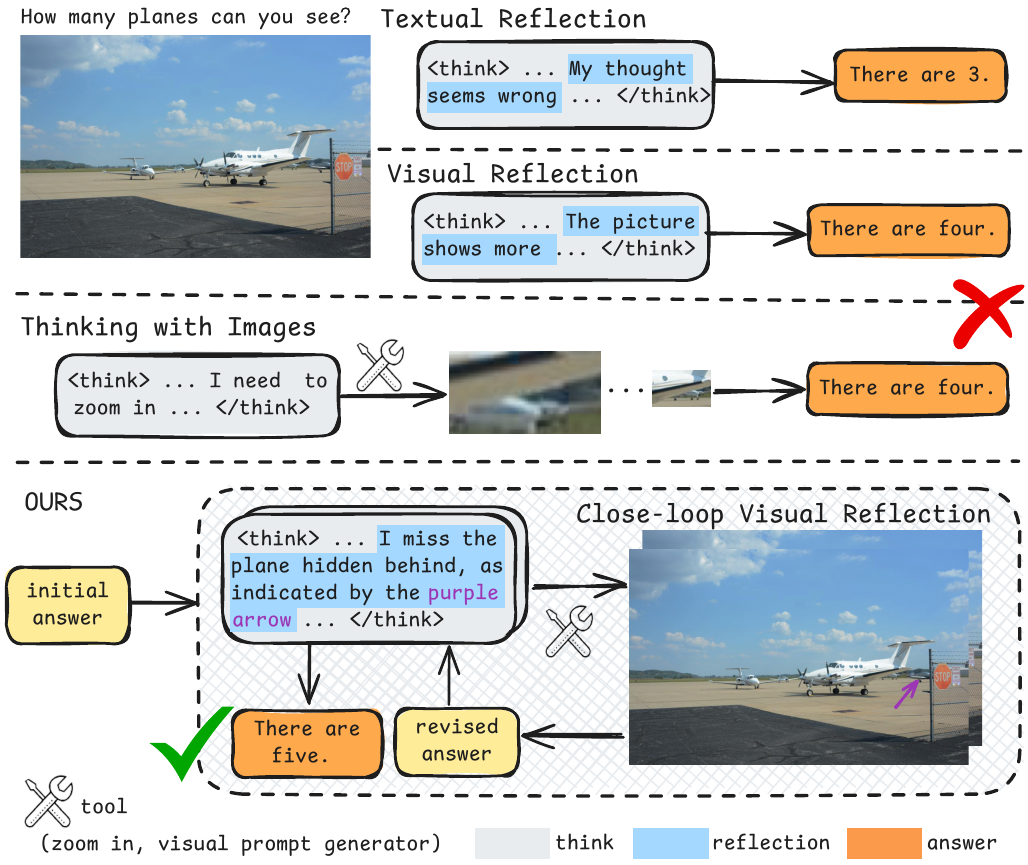}
    \caption{Comparison of existing multimodal reasoning paradigms. MIRROR explicitly seeks visual evidence to achieve closed-loop visually grounded verification.}
    \label{fig:teaser}
    \vspace{-10pt}
\end{figure}

Recently, some approaches have sought to incorporate visual information into self-reflection~\cite{look-back,Reflection-V}.
Despite improvements, these methods rely on implicit visual refocusing through attention mechanisms rather than explicit visual verification.
Crucially, while these works identify that visual attention progressively diminishes as token sequences grow, their implicit feature-guided designs inherently make it difficult to efficiently integrate visual information in the self-reflection process.
This bottleneck is further reinforced by the recent analysis~\cite{aha}, which demonstrates that inference-time self-verification behaviors often fail to yield reliable reasoning improvements.
As a result, models often produce response revisions that remain unsupported by the underlying visual evidence---a limitation we formalize as modality disconnect in self-reflection: the model refines its answer without effectively referring to visual evidence. 
Therefore, self-reflection should explicitly ground reflective reasoning in verified visual evidence.

The paradigm of ``Thinking with Images''~\cite{thinkingwithimages} provides a promising foundation for closing the gap between reflective reasoning and visual perception. By enabling region-based inspection, this approach can incorporate visual information during multi-turn reasoning. However, existing methods following this approach~\cite{openthinkimg,chainoffocus} typically operate in an open-loop manner---invoking visual operations based on the input question, rather than in response to incorrect reasoning steps. As a result, this can lead to redundant or missed visual verification.

In this paper, we propose a framework for Multimodal Iterative Reasoning via Reflection On visual Regions (MIRROR). 
It addresses modality disconnect in self-reflection by making visual evidence an explicit part of the reflection loop.
As shown in \cref{fig:teaser}, MIRROR iteratively performs four stages: response generation, error identification, visual verification, and answer revision. At each turn, the model generates a visual prompt to inspect relevant visual evidence for response uncertainties and revises its response accordingly, coupling reflective reasoning with explicit visual verification.
We further construct ReflectV, a visual reflection dataset of approximately 24k samples, ranging from general VQA to complex reasoning tasks, to endow the model with this capability.
The dataset combines internal reflection with precise visual cues to teach the model how to perform visual reflection.
We fine-tune Qwen2.5-VL (3B/7B) and Qwen3-VL-8B \cite{qwen2.5vl,qwen3vl} on ReflectV, and evaluate their performance on diverse benchmarks.
Experimental results show that MIRROR improves performance and reduces hallucinations, demonstrating the importance of transforming self-reflection from open-loop textual revision into closed-loop visually grounded verification.

In summary, our contributions are as follows:

\begin{itemize}[itemsep=2pt, parsep=0pt, topsep=2pt]

    \item We propose MIRROR, a closed-loop visual reflection framework that bridges modality disconnect in self-reflection by transforming open-loop textual revision into visually grounded verification.
    \item We construct ReflectV, a high-quality dataset of about 24k samples through a multi-agent pipeline. The dataset explicitly models trajectories of error detection, visual verification, and answer correction, teaching the model when to doubt itself and how to seek visually grounded evidence.
    \item We fine-tune multiple VLM architectures (Qwen2.5-VL 3B/7B and Qwen3-VL-8B) on our dataset to obtain the MIRROR models, and extensive evaluations demonstrate the efficacy and generalizability of our framework.

\end{itemize}

\begin{figure*}[ht]
    \centering
    \includegraphics[width=\textwidth]{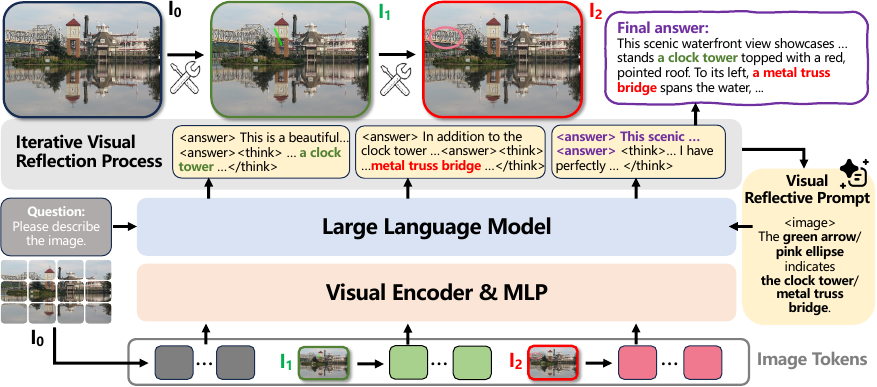}
    \caption{MIRROR performs closed-loop: it iteratively generates an initial response, identifies response uncertainties, invokes the visual prompt generator for verification, and revises based on the rendered visual evidence.}
    \label{fig:method}
    \vspace{-10pt}
\end{figure*}

\section{Related Work}



\subsection{Multimodal Reasoning in VLMs}
Multimodal reasoning in VLMs has shifted from visual question answering toward advanced Chain-of-Thought (CoT) mechanisms~\cite{cot,llava-cot} that perform reasoning through step-by-step intermediate inferences.
Research in visual grounding has employed visual prompts---such as bounding boxes and points---to enhance referential precision.
Models like Shikra \cite{shikra} and SoM-GPT4V \cite{som} demonstrate that explicit visual markers can significantly help in capturing fine-grained details.
Beyond visual grounding, recent works have explored a broader \emph{``Thinking with Images"} paradigm \cite{thinkingwithimages,openthinkimg,chainoffocus}, where VLMs actively attend to visual evidence during reasoning to acquire finer-grained visual signals for verification.
However, prior works lack a reflection-driven control signal and rarely leverage visual tools to verify response uncertainties against visual evidence.

\subsection{Reflection in Vision-Language Models}
Reflection in VLMs aims to improve reasoning reliability by allowing models to critique and revise their own predictions. Recent methods explore this capability through textual self-correction, addressing issues such as logical inconsistency, insufficient utilization of visual information, and hallucination.
For instance, Critique Fine-Tuning \cite{cft} integrates explicit critique steps into the training objective, while methods like FIRE \cite{fire} employ ``response-feedback-refinement" triplets to enable models to iteratively refine answers based on feedback signals.
SRPO \cite{srpo} introduces the <reflection> token and utilizes GRPO \cite{deepseek-math} to optimize the model's reflective capability.
While these methods strengthen logical reasoning, they often suffer from text-level hallucinations, motivating recent works to shift toward visual reflection.
VL-Rethinker \cite{vl-thinker} adopts selective sample replay for effective data filtering and forced rethinking strategies, alleviating the diminishing reflection advantage.
Look-Back \cite{look-back} designs a special token <back> to guide the model to re-attend to image information.
Reflection-V \cite{Reflection-V} leverages a supervised fine-tuning phase to align reasoning with granular visual details, followed by reinforcement learning with attention-aware rewards to reinforce the model's focus on critical regions.
Additionally, some other studies attempt to improve reasoning performance through multi-agent collaboration (e.g., Mulberry \cite{mulberry}) or tool-chain expansion (e.g., Thyme \cite{thyme}).

Unlike most existing approaches, which often fail to re-verify visual evidence explicitly, the proposed MIRROR enables visually grounded reflection. By integrating visual tools, our framework transforms reflection from implicit visual refocusing into an active verification process, where the model explicitly reflects and checks specific image regions to support answer revision with visual evidence.

\section{Method}
\label{sec:method}

MIRROR is shown in \cref{fig:method}, which transforms one-pass VLM inference into a closed-loop verification process:
(i) generate an initial response,
(ii) identify uncertainties within the response,
(iii) verify visual evidence by invoking a visual prompt generator to mark task-relevant regions,
and (iv) refine the final answer based on the updated image.

\begin{figure*}[t]
    \centering
    \includegraphics[scale=0.8]{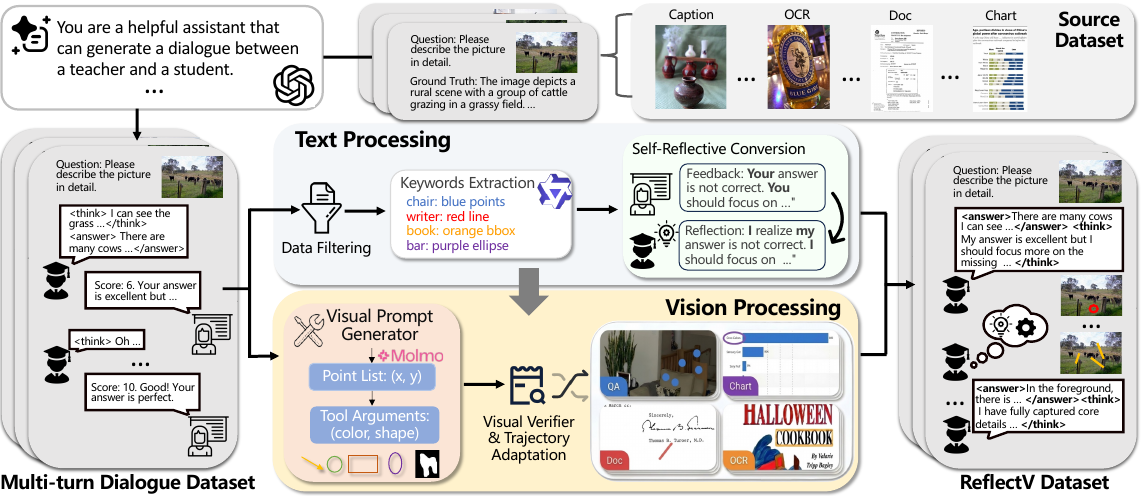}
    \caption{Overview of the ReflectV construction pipeline. Multi-turn reflective trajectories are constructed from diverse multimodal datasets and further processed into visually grounded verification trajectories.
}
    \label{fig:dataset_process}
    \vspace{-10pt}
\end{figure*}

\subsection{Formulation}

We formulate visual reflection as a multi-turn generation process.
Given an image $I_0$ and a user query $q$, the objective of the VLM, denoted as $\pi_\theta$, is to generate a structured reflection trajectory $\mathcal{Y} = \{y_1, y_2, \dots, y_K\}$ over $K$ interaction turns.
At each turn $k$, the model generates an output tuple $y_k = (a_k, r_k, v_k)$, where $a_k$ is the textual response, $r_k$ specifies the reflection content, and $v_k$ denotes the verification specification.
Unlike CoT reasoning, which operates on a fixed visual input, our framework actively verifies visual evidence during reflection.
The visual verification specification $v_k$ triggers a visual prompt generator $\mathcal{G}$, yielding an updated image $I_k = \mathcal{G}(I_0, v_k)$ that highlights specific regions for the subsequent turn.
For the $(k+1)$-th turn, the model takes the updated image $I_k$ and the interaction history $h_{<k+1} = \{a_j, r_j, v_j\}_{j=1}^{k}$  as input to generate the next response tuple $y_{k+1}$.
This process can be formulated as
\begin{equation}
    y_{k+1} \sim \pi_\theta(y_{k+1} \mid I_k, q, h_{<k+1}),
\end{equation}
where $y_{k+1} = (a_{k+1}, r_{k+1}, v_{k+1})$ represents the newly generated output tuple.
This loop continues until the model determines that no further visual verification is needed.

\subsection{Closed-Loop Visual Verification}
\label{method:tool}
A key innovation of MIRROR is the integration of a visual prompt generator $\mathcal{G}$ for visual reflection.
This module allows the model to actively ``verify'' its response by grounding textual uncertainties into visual cues.
The workflow proceeds as follows.

\noindent\textbf{Verification specification generation.} When the model identifies content that requires further visual verification, it initiates a specification enclosed within a special token <tool\_call> structure $v_k$ =
{
    \setlength{\abovedisplayskip}{5pt}
    \setlength{\belowdisplayskip}{5pt}
    \begin{equation}
        \resizebox{0.8\hsize}{!}{
            $\begin{aligned}
                & \texttt{<tool\_call>} \{ \\
                & \quad \texttt{name}: \texttt{"Visual Prompt Generator"}, \\
                & \quad \texttt{flag}: \texttt{true/false} \\
                & \quad \texttt{anchor}: \texttt{"keywords..."} \\
                & \quad \texttt{args}: \{ \texttt{color}: \texttt{"red"}, \texttt{mark\_type}: \texttt{"point"} \} \\
                & \} \texttt{</tool\_call>} ,
            \end{aligned}$
        }
    \end{equation}
}

\noindent where \texttt{name} invokes the visual module, \texttt{anchor} specifies the textual region derived from reflection content for localization, and \texttt{args} defines the rendering attributes for visual prompts. The \texttt{flag} field controls whether visual tools are invoked at each turn: when set to \texttt{true}, the model requests grounded visual prompts to verify its response, and when \texttt{false}, the current answer is accepted as the final output without further visual verification. 


\noindent\textbf{Visual grounding.} Upon receiving $v_k$, the generator $\mathcal{G}$ localizes the anchor, predicts its image coordinates, and overlays the requested marker (e.g., "red point") according to the arguments, producing an updated image $I_k = \mathcal{G}(I_0, v_k)$.

\noindent\textbf{Reasoning refinement.} The resulting image $I_k$, now explicitly highlighting the neglected or misinterpreted regions, is fed back into the VLM.
This process forms an iterative feedback loop, forcing the model to re-attend to the updated image ($I_k$) alongside the query history before generating the next reasoning step.

\subsection{Training for Reflective Reasoning}
We utilize a visual reflection dataset $\mathbb{D}$ (detailed in \cref{sec:dataset}) to train the VLM via supervised fine-tuning.
The training objective is to optimize the model's ability to iteratively execute a closed-loop visual reflection process.
We formulate the training objective as a multi-turn auto-regressive loss.
Formally, for a training sample $(I_0, q, \mathcal{Y})$, we maximize the likelihood of the sequence over $K$ turns:
\begin{equation}
\resizebox{\linewidth}{!}{%
$ \displaystyle
\min_{\theta} \mathbb{E}_{(I_0, q, \mathcal{Y}) \sim \mathbb{D}} \bigg[ - \sum_{k=1}^{K} \log \pi_\theta \big(a_k, r_k, v_k \mid  I_{k-1}, q, h_{<k} \big) \bigg],
$
}
\end{equation}
where $h_{<k} = \{a_j, r_j, v_j\}_{j=1}^{k-1}$ represents the input history, and $I_{k-1}$ denotes the updated image processed by the visual prompt generator in the previous turn.
This objective jointly supervises answer revision ($a_k$), reflective reasoning ($r_k$), and visual verification specification generation ($v_k$).

\section{Visual Reflection Dataset Construction}
\label{sec:dataset}

We construct ReflectV as illustrated in \cref{fig:dataset_process}. The construction process consists of three stages: (i) multi-turn dataset construction, which generates and filters reflective reasoning chains, (ii) visual evidence construction, which converts textual reflection into visual prompts and visually grounded verification trajectories, and (iii) trajectory refinement and adaptation, which produces mixed multi-turn and single-turn supervision for both iterative correction and efficient one-pass inference. Additional details on data sources, pipeline implementation, and data filtering strategy ablations are provided in Appendix~\ref{sec:supp_data}.

\subsection{Multi-Turn Dataset Construction}
\label{sec4:data_filter}
We adopt the data curation pipeline proposed in FIRE~\cite{fire} to construct reflective reasoning trajectories for visually grounded reflection.
Instead of relying solely on existing conversational datasets, we transform QA pairs into multi-turn dialogues.
Specifically, an advanced VLM simulates a ``student-teacher'' interaction based on ground-truth annotations, where the ``student'' produces intermediate responses containing potential uncertainties, and the ``teacher'' provides corrective feedback and a score $s_t$ until the correct answer is given.
This process converts QA samples into rich trajectories of error identification and answer revision.

To ensure the quality of the generated trajectories, we further apply a filtering strategy based on the teacher score $s_t$.
A trajectory is preserved only if its scores strictly increase across turns ($s_{t+1} > s_t$), the dialogue terminates with a maximum final score ($s_{\text{final}} = 10$), and the final response is semantically aligned with the ground-truth answer, preventing false positives caused by verifier misjudgment.

\subsection{Visual Evidence Construction}

To explicitly connect reflective reasoning with visual evidence, we convert textual reflection into visually grounded verification trajectories.

First, we extract verification keywords from the reflection content and use them as tool arguments for visual prompting (\cref{method:tool}). For general VQA tasks, entity-centric keywords are directly parsed from the reflection.
For dense perception tasks such as OCR and chart understanding, where reflections often lack spatial specificity, we instead derive verification keywords from question entities.
To further strengthen alignment between reflection and visual evidence, visual attribute descriptions (e.g., ``as indicated by the red point'') are injected back into the reflection text.

Second, we employ a self-reflection conversion mechanism to convert external corrective feedback into first-person reflective reasoning.
Concretely, an LLM rewrites teacher feedback (e.g., ``Your answer is incorrect'') into self-reflective thoughts (e.g., ``Upon closer inspection, I realize my previous answer was incorrect'').
This conversion process teaches the model how to identify potentially incorrect response elements and initiate visual verification during reflection.

\subsection{Trajectory Refinement and Adaptation}
\label{sec4:visual_verification}

After visual prompting is applied, we further refine the trajectories to ensure reliable visual grounding and balanced supervision.

First, we perform visual consistency filtering using a visual verifier to assess the semantic alignment between generated visual prompts and reflection content.
This step removes instances with ambiguous or incorrectly grounded visual cues.

Second, we adopt a mixed trajectory adaptation strategy.
Verified reflective trajectories are preserved as multi-turn supervision for iterative visual verification, while failed or redundant trajectories are simplified into direct QA pairs $(I_0, q \rightarrow a_{\text{final}})$.
This mixed supervision strategy improves both closed-loop reflective reasoning and efficient one-pass inference while reducing over-reliance on unnecessarily long correction chains.

\begin{table*}[t]
    \centering
    \caption{Performance on General and OCR \& Document benchmarks.
The best and second-best results are highlighted in \textbf{bold} and \underline{underlined}. MIRROR (w/o loop) is an open-loop variant without reflective loop.}
    \label{tab:general_ocr}
    \renewcommand{\arraystretch}{1.2}
    \setlength{\tabcolsep}{8pt}
    \resizebox{\textwidth}{!}{
        \begin{tabular}{l ccc ccc}
            \toprule
            \textbf{Model} & \multicolumn{3}{c}{\textbf{General Capabilities}} & \multicolumn{3}{c}{\textbf{OCR \& Document}} \\
            \cmidrule(lr){2-4} \cmidrule(lr){5-7}
             & \textbf{MM-Vet} & \textbf{MMStar} & \textbf{SeedBench-2-Plus} & \textbf{TextVQA-Val} & \textbf{OCRBench} & \textbf{ChartQA-Test} \\
            \midrule
            LLaVA-OneVision-7B & 48.80 & 61.70 & -- & 76.10 & 62.10 & 80.00 \\
            InternVL3-2B & 54.95 & 60.70 & 64.95 & 77.00 & 82.20 & 76.08 \\
            InternVL3-8B & \underline{64.27} & 61.50 & 69.61 & 80.51 & 85.00 & 79.64 \\
            Qwen2.5-VL-3B & 47.39 & 55.87 & 68.81 & 79.12 & 82.60 & 83.20 \\
            Qwen2.5-VL-7B & 56.60 & 61.21 & \underline{70.88} & 84.90 & 83.20 & 86.08 \\
            \midrule
            \rowcolor{gray!15} \textbf{MIRROR}(w/o loop) & 59.91 & \underline{62.80} & 70.36 & \underline{85.37} & \underline{88.30} & \underline{86.56} \\
            \rowcolor{gray!15} \textbf{MIRROR}(ours) & \textbf{66.70} & \textbf{73.33} & \textbf{76.86} & \textbf{86.62} & \textbf{92.00} & \textbf{87.92} \\
            \bottomrule
        \end{tabular}
    }
    \vspace{-5pt}
\end{table*}

\begin{table*}[t]
    \centering
    \caption{Performance comparison on Hallucination, Fine-grained Perception, and Math benchmarks.
The best and second-best results are highlighted in \textbf{bold} and \underline{underlined}, respectively. MME-RW indicates MME-RealWorld-Lite. }
    \label{tab:hallucination_reasoning}
    \renewcommand{\arraystretch}{1}
    \setlength{\tabcolsep}{8pt}
    \resizebox{\textwidth}{!}{
        \small
        \begin{tabular}{l cc ccc c}
            \toprule
            \textbf{Model} & \multicolumn{2}{c}{\textbf{Hallucination}} & \multicolumn{3}{c}{\textbf{Fine-grained Perception}} & \multicolumn{1}{c}{\textbf{Math}} \\
            \cmidrule(lr){2-3} \cmidrule(lr){4-6} \cmidrule(lr){7-7}
             & \textbf{POPE} & \textbf{HallusionBench} & \textbf{HRBench-4K} & \textbf{MME-RW} & \textbf{VStarBench} & \textbf{MathVision} \\
            \midrule
            LLaVA-OneVision-7B & 78.10 & 31.60 & 63.00 & -- & 72.30 & 18.30 \\
            InternVL3-2B & 89.60 & 42.50 & 61.75 & 43.88 & 68.59 & 21.71 \\
            InternVL3-8B & \underline{90.37} & 49.90 & \underline{70.00} & \underline{48.83} & 68.06 & 20.39 \\
            Qwen2.5-VL-3B & 86.21 & 63.09 & 50.25 & 42.15 & 72.77 & 25.66 \\
            Qwen2.5-VL-7B & 86.45 & \underline{68.66} & 68.87 & 44.29 & 75.39 & 23.36 \\
            \midrule
            \rowcolor{gray!15} \textbf{MIRROR}(w/o loop) & 87.95 & 68.24 & 69.13 & 46.01 & \underline{76.44} & \underline{27.30} \\
            \rowcolor{gray!15} \textbf{MIRROR}(ours) & \textbf{94.42} & \textbf{82.02} & \textbf{72.88} & \textbf{51.49} & \textbf{83.77} & \textbf{28.29} \\
            \bottomrule
        \end{tabular}
    }
    \vspace{-10pt}
\end{table*}

\begin{figure*}[t]
    \centering
    \includegraphics[width=\textwidth]{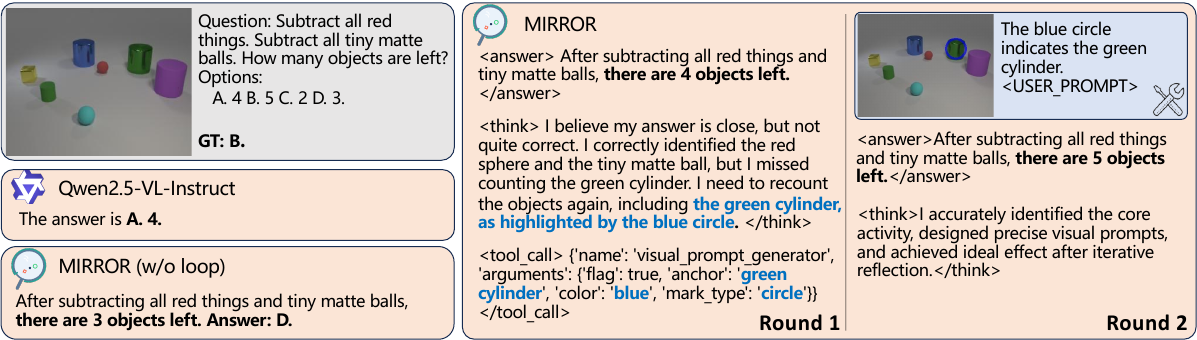}
    \caption{Qualitative examples of iterative visual reflection.
    Compared to Qwen2.5-VL and MIRROR (w/o loop) (without reflective loop), MIRROR successfully corrects errors via visually grounded verification.}
    \label{fig5:real_case}
    \vspace{-10pt}
\end{figure*}



\section{Experiment}


\subsection{Experimental Setting}

We conduct experiments across 12 diverse benchmarks. For general capabilities, we employ MM-Vet \cite{mmvet}, MMStar \cite{mmstar}, and SeedBench-2-Plus \cite{seedbench2}.
In the domain of OCR and document understanding, we utilize TextVQA \cite{textvqa}, OCRBench \cite{ocrbench}, and ChartQA \cite{chartqa}.
To assess the models' robustness against hallucinations, we include POPE \cite{pope} and HallusionBench \cite{hallusionbench}.
Furthermore, we evaluate fine-grained perception using HRBench \cite{hrbench}, MME-RealWorld-Lite \cite{mmerealworld}, and VStarBench \cite{vstar}, alongside MathVision \cite{mathvision} for mathematical reasoning.
We employ VLMEvalKit \cite{vlmevalkit} to evaluate models on these benchmarks, and use GPT-5-mini \cite{gpt5} as the judge to evaluate the correctness of the model's responses for datasets if needed.

\begin{figure*}[t]
    \centering
    \includegraphics[width=\textwidth]{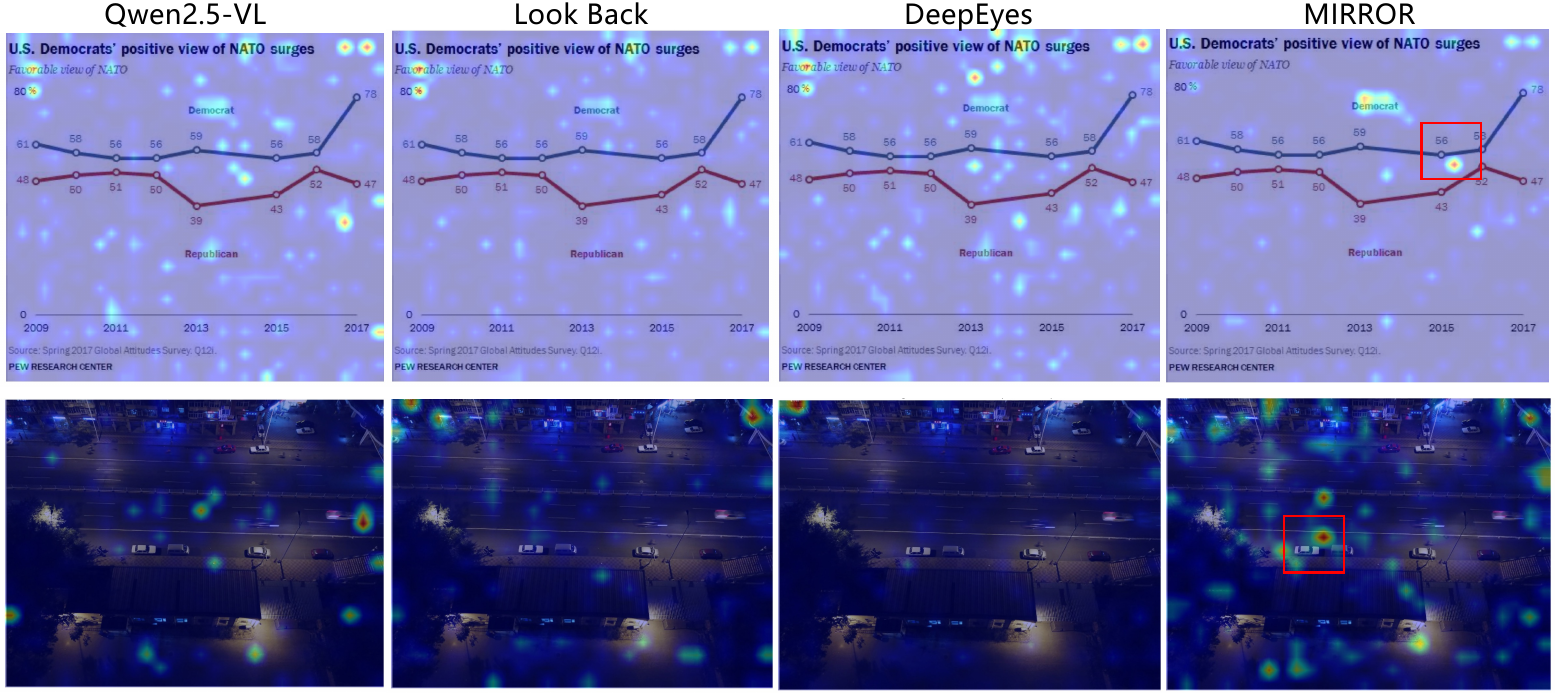}
    \caption{Visualization of per-token attention maps in the first reasoning round (before tool invocation). 
    Correct target regions (the specific value ``56'' and the ``van'') are highlighted with red boxes in MIRROR's attention maps.
    }
    \label{fig:attention}
    \vspace{-10pt}
\end{figure*}

\subsection{Comparison with Base Models}

We compare MIRROR with strong VLMs, including Qwen2.5-VL \cite{qwen2.5vl}, InternVL3 \cite{internvl3}, LLaVA-OneVision \cite{llavaonevision}.
As shown in \cref{tab:general_ocr,tab:hallucination_reasoning}, MIRROR consistently outperforms the base model and other strong baselines, while MIRROR (w/o loop) also achieves notable performance gains. We also provide more qualitative visualization in Appendix~\ref{sec:supp_vis} and results of other model sizes and architectures in Section \ref{sec:ablation}.

\noindent\textbf{General Capabilities and Math Reasoning.}
On benchmarks evaluating broad multimodal skills and mathematical reasoning, MIRROR demonstrates significant gains over the base models.
As illustrated in \cref{fig5:real_case}, other models often fail in tasks requiring multi-step logic due to attention drift.
In contrast, MIRROR actively identifies and resolves logical discrepancies during complex problem-solving. Notably, despite the absence of mathematical data, MIRROR still achieves notable enhancements in
foundational mathematical logic.

\noindent\textbf{OCR \& Document Understanding.}
On text-intensive tasks, MIRROR achieves the best performance among the compared models.
MIRROR helps the model explicitly revisit neglected visual details, overcoming the limitations inherent in one-pass inference. We provide a representative correction case in \cref{fig:reasoning_case1}.

\noindent\textbf{Hallucination and Fine-grained Perception.}
Crucially, MIRROR excels in reducing hallucinations (e.g., +13.36\% on HallusionBench) and enhancing fine-grained perception.
By anchoring intermediate reasoning in verifiable visual evidence, our method imposes a strong constraint on the trajectory.

\subsection{Comparison with Reasoning Models}

\textbf{Implementation Setup.} To evaluate the effectiveness of our closed-loop visual reflection mechanism, we compare MIRROR with representative methods from three reasoning paradigms: textual reflection, visual reflection, and Thinking with Images. Specifically, we include VL-Rethinker~\cite{vl-thinker} for textual reflection and Look-Back~\cite{look-back} for visual reflection, reporting both its solution-level and semantic-level variants. For Thinking with Images, we compare against PixelReasoner~\cite{pixelreasoner}, DeepEyes~\cite{deepeyes}, and Adaptive-CoF~\cite{chainoffocus}. We select one representative benchmark from each of four major categories and include the SFT variants of several models to ensure a fair comparison.

\begin{table}[t]
    \centering
    \begin{threeparttable}
        \caption{Performance comparison with SOTA reasoning methods, all fine-tuned on Qwen2.5-VL-7B. Best and second-best results are \textbf{bold} and \underline{underlined}.}
        \label{tab:reasoning_comparison}
        \renewcommand{\arraystretch}{1}
        \setlength{\tabcolsep}{1pt}
        \footnotesize
        \begin{tabular}{l cccc}
            \toprule
            \textbf{Method} & \textbf{OCRBench} & \textbf{POPE} & \textbf{MME-RW} & \textbf{MM-Vet} \\
            \midrule
            \rowcolor{gray!10} \multicolumn{5}{l}{\textit{Text Reflection}} \\
            VL-Rethinker & 85.40 & 84.19 & 47.21 & 56.19 \\
            \midrule
            \rowcolor{gray!10} \multicolumn{5}{l}{\textit{Visual Reflection}} \\
            LookBack (Solution) & 87.50 & 88.20 & 49.80 & 63.50 \\
            LookBack (Semantic) & \underline{88.60} & \underline{89.80} & 50.40 & 65.10 \\
            \midrule
            \rowcolor{gray!10} \multicolumn{5}{l}{\textit{Thinking with Images}} \\
            PixelReasoner-SFT & 76.35 & 80.01 & 44.73 & 47.68 \\
            PixelReasoner & 82.10 & 86.03 & 49.70 & 52.98 \\
            DeepEyes & 88.10 & 87.70 & 49.50 & 60.28 \\
            Adaptive-CoF-SFT & 85.62 & 82.53 & 50.10 & 62.73 \\
            Adaptive-CoF & 86.00 & 89.30 & \underline{50.90} & \underline{66.21} \\
            \midrule
            \rowcolor{gray!10} \textbf{MIRROR} & \textbf{92.00} & \textbf{94.42} & \textbf{51.49} & \textbf{66.70} \\
            \bottomrule
        \end{tabular}
    \end{threeparttable}
    \vspace{-20pt}
\end{table}

As shown in \cref{tab:reasoning_comparison}, MIRROR achieves superior performance by addressing the inherent limitations of existing paradigms.
First, pure text-based reflection suffers from hallucinations. For instance, VL-Rethinker scores low on POPE (84.19), confirming that self-correction without visual information integration is insufficient.
Second, while ``Thinking with Images'' and ``Visual Reflection'' methods introduce active visual tools or latent refocusing, they largely operate in an open-loop manner, lacking an error-correction mechanism to rectify perceptual failures.
In contrast, MIRROR incorporates a closed-loop visually grounded verification process, consequently surpassing the strongest baselines across all metrics. We also provide more efficiency analysis and visualization in Appendix~\ref{sec:supp_eff_rea}.

\noindent\textbf{Attention Visualization.}
To understand why MIRROR achieves stronger reasoning, we visualize the average per-token attention maps in \cref{fig:attention}. To isolate the model's intrinsic reflection capability, all visualizations are captured from the first reasoning round before any tool invocation. As shown in \cref{fig:attention}, baseline methods exhibit scattered or misaligned attention patterns, struggling to focus on task-critical regions. In contrast, MIRROR consistently demonstrates concentrated attention on correct areas across both reasoning-intensive and perceptual-grounding tasks. This suggests that the reflective training paradigm improves the model’s ability to focus on task-relevant visual regions even before external tool invocation.

\subsection{Ablation Experiments}
\label{sec:ablation}

\noindent\textbf{Impact of Reflective Training Paradigm.}
To disentangle the benefits of our closed-loop reflection mechanism from the intrinsic quality of the domain-specific knowledge in ReflectV, we compare MIRROR against vanilla-SFT variants fine-tuned on the same dataset but reformatted into QA pairs.
As shown in \cref{tab:ablation_paradigm}, vanilla-SFT variants outperform the base models due to domain adaptation, yet MIRROR achieves substantially higher gains across all model sizes and architectures, validating the effectiveness of our reflective training paradigm.

\begin{table}[htbp]
    \centering
    \caption{Ablation study on training paradigms, model sizes, and architectures. Best and second-best results are \textbf{bold} and \underline{underlined}.}
    \label{tab:ablation_paradigm}
    \renewcommand{\arraystretch}{1.1}
    \footnotesize
    \setlength{\tabcolsep}{2pt}
    \begin{tabular}{l cccc}
        \toprule
        \textbf{Method} & \textbf{OCRBench} & \textbf{POPE} & \textbf{VStarBench} & \textbf{MM-Vet} \\
        \midrule
        Qwen2.5-VL-3B & \underline{82.60} & 86.20 & 72.77 & 47.39 \\
        +vanilla-SFT & 81.60 & \underline{86.31} & \underline{75.92} & \underline{56.33} \\
        \textbf{MIRROR-3B} & \textbf{85.40} & \textbf{87.74} & \textbf{80.63} & \textbf{60.60} \\
        \midrule
        Qwen2.5-VL-7B & 83.20 & 86.45 & 75.39 & 56.60 \\
        +vanilla-SFT & \underline{86.90} & \underline{86.99} & \underline{78.01} & \underline{61.01} \\
        \textbf{MIRROR-7B} & \textbf{92.00} & \textbf{94.42} & \textbf{83.77} & \textbf{66.70} \\
        \midrule
        Qwen3-VL-8B & 90.50 & 88.07 & 74.35 & 67.00 \\
        +vanilla-SFT & \underline{90.80} & \underline{88.46} & \underline{75.23} & \underline{67.52} \\
        \textbf{MIRROR-8B} & \textbf{93.30} & \textbf{95.43} & \textbf{86.83} & \textbf{79.56} \\
        \bottomrule
    \end{tabular}
\end{table}

\noindent\textbf{Scalability across Model Sizes and Architectures.}
We further investigate the universality of our approach by applying MIRROR to different model sizes and architectures.
As detailed in \cref{tab:ablation_paradigm}, MIRROR achieves consistent gains across all configurations: for 3B/7B/8B variants, it improves over the base model by +13.21/+10.10/+12.6 points on MM-Vet, and by +2.80/+8.80/+2.80 on OCRBench.
Notably, MIRROR-8B achieves the highest MM-Vet score, demonstrating that our closed-loop visually grounded verification mechanism effectively transfers across different VLM architectures (Qwen2.5-VL and Qwen3-VL).
This confirms that our framework is both model-agnostic and architecture-agnostic, generalizing well to more VLM backbones with enhanced native reasoning capabilities.

\noindent\textbf{Trajectory Adaptation.}
To evaluate the effectiveness of our trajectory adaptation strategy, we investigate the impact of $\rho$, defined as the ratio of multi-turn reflective chains, as shown in \cref{tab3:ablation_ratio}.
Training exclusively on multi-turn reflective data ($\rho=1.0$) leads to a significant degradation compared to the mixed strategy.

\begin{table}[h]
    \centering
    \caption{Ablation study regarding multi-turn sample ratio $\rho$ for trajectory adaptation on MM-Vet. Best and second-best results are \textbf{bold} and \underline{underlined}.}
    \label{tab3:ablation_ratio}
    \renewcommand{\arraystretch}{1.1}
    \footnotesize
    \setlength{\tabcolsep}{4pt}
    \begin{tabular}{lcccccc}
        \toprule
        \textbf{Ratio $\rho$} & \textbf{OCR} & \textbf{Math} & \textbf{Spat} & \textbf{Gen} & \textbf{Rec} & \textbf{Avg.} \\
        \midrule
        $\rho=0.5$ & 61.85 & 51.54 & 59.73 & 45.75 & 55.61 & 55.87 \\
        $\rho=0.6$ & 64.81 & 56.15 & \underline{62.13} & \underline{56.75} & \underline{61.28} & \underline{60.32} \\
        $\rho=1.0$ & \underline{65.19} & \underline{56.54} & 61.47 & 54.12 & 59.04 & 58.81 \\
        \midrule
        \rowcolor{gray!15}
        \textbf{$\rho=0.75$} & \textbf{69.90} & \textbf{60.77} & \textbf{70.80} & \textbf{61.62} & \textbf{64.40} & \textbf{66.70} \\
        \bottomrule
    \end{tabular}
    \vspace{-10pt}
\end{table}

\begin{table}[h]
    \centering
    \caption{Ablation study of the closed-loop reflection mechanism on the MM-Vet benchmark (Average Score). \textbf{Bold} and \underline{underlined} indicate the best and second-best performance, respectively.}
    \label{tab:component_ablation1}
    \renewcommand{\arraystretch}{1.05}
    \footnotesize
    \setlength{\tabcolsep}{6pt}
    \begin{tabular}{l c c c}
        \toprule
        \textbf{Configuration} & \textbf{Loop} & \textbf{Tools} & \textbf{MM-Vet} \\
        \midrule
        Qwen2.5-VL-7B         & -          & -          & 56.60 \\
        Qwen2.5-VL-7B-Tools   & -          & \checkmark & 57.90 \\
        Qwen2.5-VL-7B-MIRROR  & \checkmark & \checkmark & 57.10 \\
        MIRROR (w/o loop)     & -          & -          & 59.91 \\
        MIRROR-Tools          & -          & \checkmark & \underline{62.70} \\
        \midrule
        \textbf{MIRROR (ours)}& \textbf{\checkmark} & \textbf{\checkmark} & \textbf{66.70} \\
        \bottomrule
    \end{tabular}
    \vspace{-15pt}
\end{table}

\noindent\textbf{Effectiveness of Closed-loop Reflection.} As illustrated in \cref{tab:component_ablation1}, we compare MIRROR with three variants to isolate the effects of visual tools and iterative closed-loop verification. Qwen2.5-VL-7B-Tools adds the same visual tools to the base model without reflective training. Qwen2.5-VL-7B-MIRROR applies the MIRROR-style loop in a few-shot setting. MIRROR-Tools disables the iterative loop. We observe two key findings.
(1) Tool access alone is insufficient. Simply equipping the base model with visual tools without closed-loop training yields only marginal gains.
(2) Verification drives the gains. Removing the iterative loop (MIRROR-Tools) causes a substantial performance drop. This confirms that our closed-loop visual verification, rather than mere tool integration, is the primary driver of MIRROR's success.
Additional efficiency comparisons with other models in Appendix~\ref{sec:supp_eff_ablation} show that this verification process incurs only modest overhead.
A detailed analysis of multi-turn outcomes is provided in Appendix~\ref{sec:reflection_outcome}.

\section{Conclusion}
In this work, we have presented MIRROR, a framework that enhances multimodal reasoning by transforming self-reflection from open-loop textual revision into closed-loop, visually grounded verification.
We construct ReflectV, a high-quality dataset designed to teach models to explicitly ground their reasoning in specific image regions.
Our experiments demonstrate that MIRROR outperforms strong baselines, validating that equipping VLMs with the agency to ``look again'' explicitly and verify visual evidence is essential for reducing hallucinations and achieving robust reasoning.

\section*{Limitations}


Despite its effectiveness, MIRROR still has limitations in both capability scope and system design. For capability scope, visual verification remains difficult in abstract domains and fine-grained compositional attribute binding, where the target evidence is hard to represent with simple spatial anchors. We provide a detailed analysis of these cases in Appendix~\ref{sec:supp_limitations}. For system design, MIRROR currently relies on an external visual prompt generator to localize anchors and render visual cues, which introduces additional parameters and system complexity beyond the backbone VLM. Future work will focus on more fine-grained verification mechanisms and explore how to internalize region-level reflection and verification through the native visual grounding capabilities of VLMs.


\bibliography{mirror_ref}

\newpage

\appendix

\begin{figure*}[htpb]
    \centering
    \includegraphics[width=\textwidth]{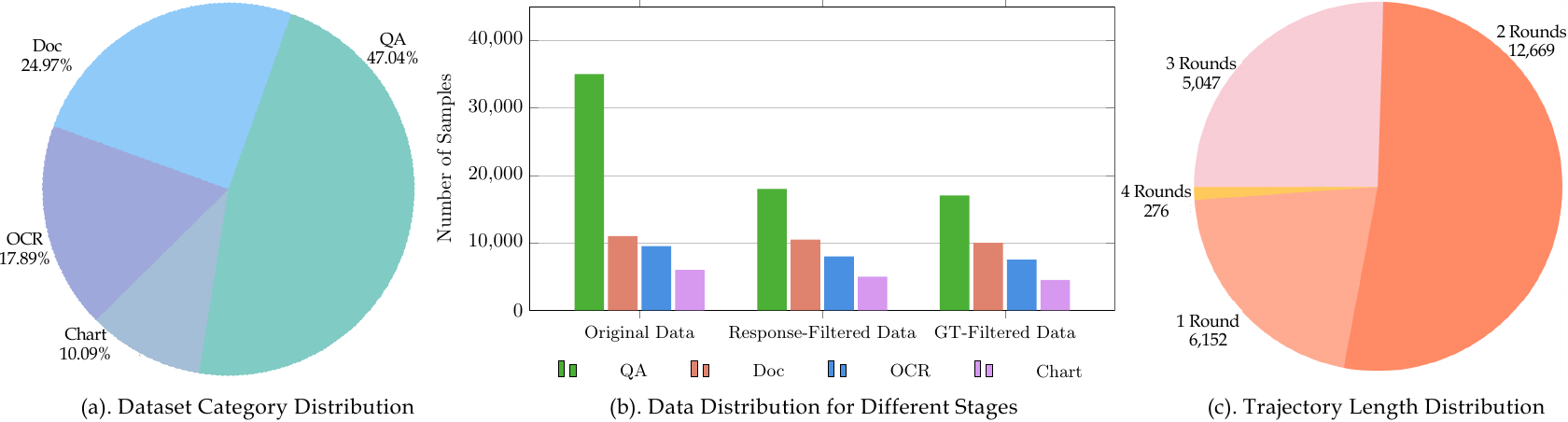} 
    \caption{\textbf{Statistics and distribution of the ReflectV dataset construction.} 
    (a). \textbf{Domain Distribution:} The composition of the raw data spans four distinct capabilities: General QA, Document Understanding (Doc), Scene Text (OCR), and Chart Reasoning. 
    (b). \textbf{Filtering Pipeline:} The data volume retention across the three construction stages (Original $\rightarrow$ Response-Filtered $\rightarrow$ GT-Filtered), illustrating the rigorous quality control process described in \cref{sec4:data_filter}. 
    (c). \textbf{Trajectory Depth:} We set the default mixing ratio to $\rho=0.75$ for our training. The distribution of samples based on the number of self-reflection rounds required for convergence illustrates the varying complexity of error correction.}
    \label{fig:data_distribution}
\end{figure*}

\section{Dataset Setting}
\label{sec:supp_data}

\subsection{Data Sources}
\label{sec:supp_data_source}
We construct a high-quality visual reflection instruction-tuning dataset comprising approximately 24k samples. The data sources are curated from diverse public benchmarks, including COCO \cite{coco}, LLaVA \cite{llava}, GQA \cite{gqa}, TextVQA \cite{textvqa}, OCRVQA \cite{ocrvqa}, DOCVQA \cite{docvqa}, DVQA \cite{dvqa}, and ChartQA \cite{chartqa}. These samples cover a wide range of visual domains and reasoning complexities, serving as the foundation for our fine-tuning process. 

For benchmarks that appear in both our training data and evaluation (TextVQA and ChartQA), we strictly use the training splits for ReflectV construction and the validation/test splits for evaluation, following the standard protocol of each benchmark. Specifically, TextVQA-Val and ChartQA-Test are held out for evaluation and never seen during training.

To ensure balanced capability acquisition, we categorize these sources into four primary domains: General QA, Scene Text (OCR), Document Understanding, and Chart Reasoning. As illustrated in \cref{fig:data_distribution}(a), this domain-specific distribution is strategically balanced to retain generalizability while boosting fine-grained perception. Furthermore, the statistics regarding the data filtering process and trajectory depth are detailed in \cref{fig:data_distribution}(b) and (c), respectively.

\subsection{Data Construction Pipeline}

Specifically, we employ Qwen2.5-7B~\cite{qwen2.5} to generate the initial textual self-reflection chains and extract keywords. Molmo-7B~\cite{molmo} is then employed to generate precise point coordinates for the mentioned keywords, which serve as prompts for {SAM 2}~\cite{sam2} (optional) to produce fine-grained segmentation masks. Finally, {Qwen2.5-VL-7B}~\cite{qwen2.5vl} acts as a visual verifier to ensure the generated cues are semantically aligned with the image content.

\subsection{Data Filtering Strategy}

To demonstrate the effectiveness of our filtering strategy and the high quality of our data, we conduct a comparative quality scoring of the filtered, unfiltered, and discarded subsets, and further validate our approach by benchmarking our model against a baseline trained on raw, unfiltered data.

First, we conduct a blind quality assessment involving both human experts and GPT-4o-mini. We randomly sample 100 trajectories from three subsets: the discarded noisy data ($D_{noise}$), the raw unfiltered data ($D_{raw}$), and our final filtered data ($D_{clean}$). Each sample is scored on a scale of 1 to 5 based on Logical Coherence and Visual Validity.

\begin{table}[h]
    \centering
    \caption{Comparative quality assessment. We evaluate samples from Discarded ($D_{noise}$), Unfiltered ($D_{raw}$), and Filtered ($D_{clean}$) subsets. The results demonstrate that our pipeline effectively isolates high-quality trajectories.}
    \label{tab:data_quality}
    \footnotesize
    \setlength{\tabcolsep}{1pt}
    \begin{tabular}{lcccccc}
        \toprule
        \multirow{2}{*}{\textbf{Dataset}} & \multicolumn{2}{c}{\textbf{Human}} & \multicolumn{2}{c}{\textbf{GPT-4o-mini}} & \multicolumn{2}{c}{\textbf{Average}} \\
        \cmidrule(lr){2-3} \cmidrule(lr){4-5} \cmidrule(lr){6-7}
         & \textbf{Logic} & \textbf{Visual} & \textbf{Logic} & \textbf{Visual} & \textbf{Logic} & \textbf{Visual} \\
        \midrule
        $D_{noise}$ & 2.14 & 1.38 & 2.30 & 1.45 & 2.22 & 1.42 \\
        $D_{raw}$  & 3.56 & 3.12 & 3.65 & 3.20 & 3.61 & 3.16 \\
        \rowcolor{gray!15} \textbf{$D_{clean}$} & \textbf{4.78} & \textbf{4.65} & \textbf{4.82} & \textbf{4.70} & \textbf{4.80} & \textbf{4.68} \\
        \bottomrule
    \end{tabular}
    \vspace{-10pt}
\end{table}

\begin{figure*}[h]
    \vspace{-10pt}
    \centering
    \includegraphics[width=0.6\textwidth]{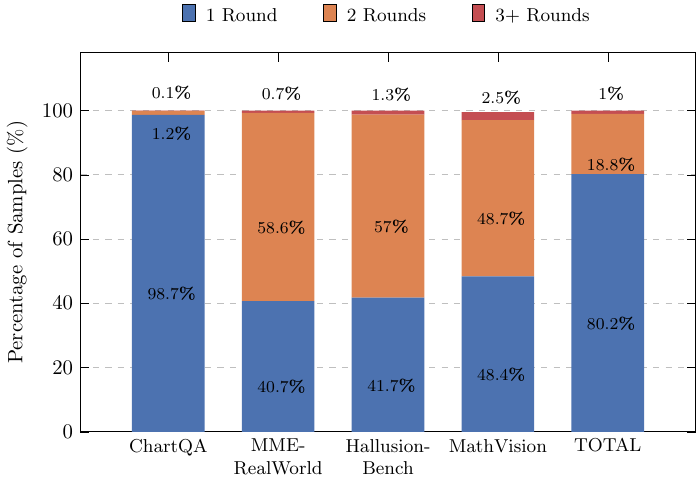}
    \caption{\textbf{Distribution of inference trajectory lengths across representative benchmarks.} The stacked bars illustrate the percentage of samples resolved within 1, 2, or 3+ rounds.}
    \label{fig:inference_rounds}
\end{figure*}

As reported in \cref{tab:data_quality}, $D_{clean}$ consistently achieves the highest scores (Avg. 4.74), significantly surpassing the raw baseline ($D_{raw}$, 3.39). Conversely, the discarded subset ($D_{noise}$) exhibits the lowest performance, particularly in visual validity (1.42). This distinct stratification confirms that our pipeline effectively identifies and prunes samples with hallucinated or ungrounded visual prompts, ensuring high-quality training data. For more visualizations of $D_{noise}$, please refer to \cref{fig:data_case1}.

\begin{figure*}[p]
    \centering
    \begin{subfigure}{\textwidth}
        \centering
        \includegraphics[width=\linewidth]{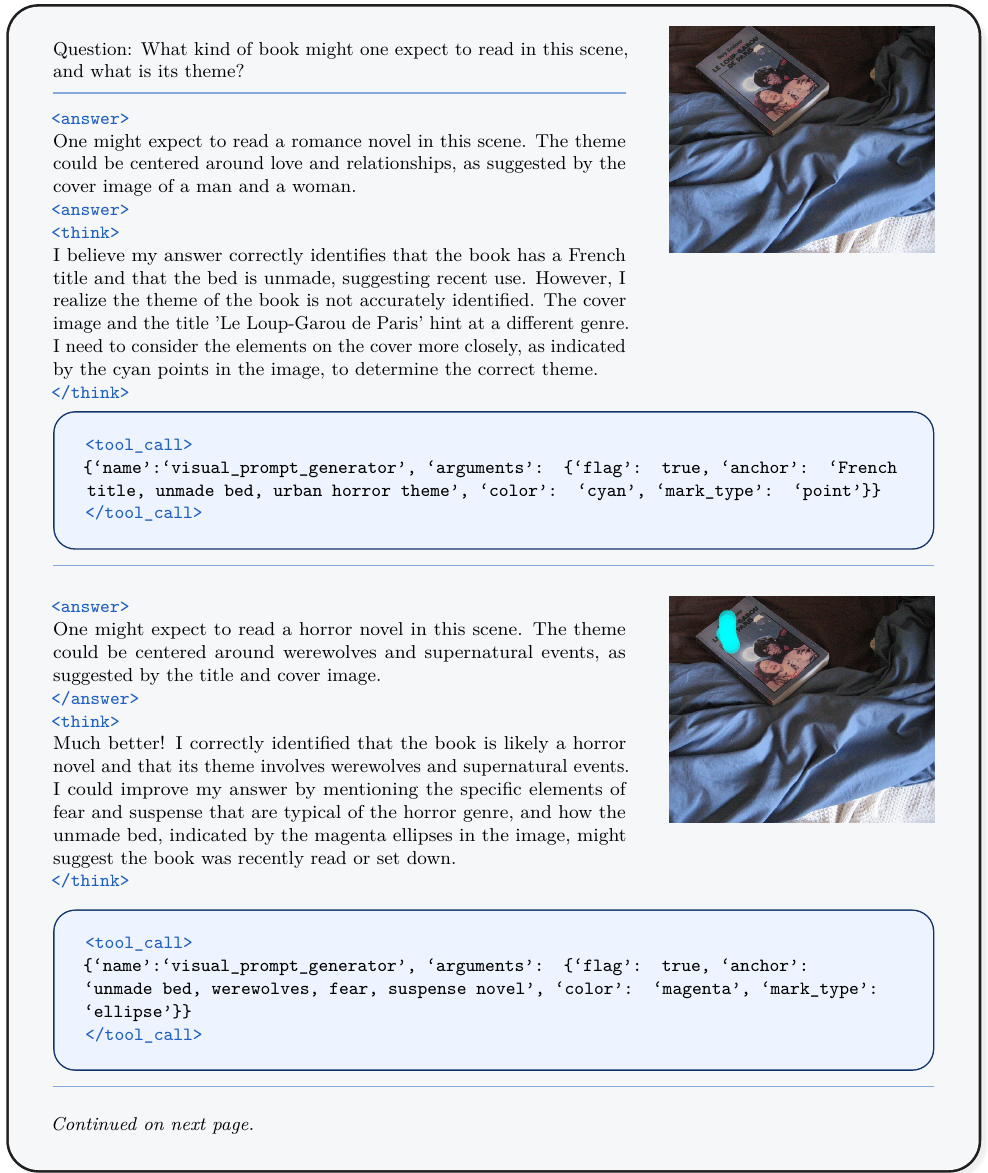}
        \caption{\textbf{Part 1 of the case from $D_{\text{noise}}$.}}
    \end{subfigure}
\end{figure*}

\begin{figure*}[t!]
    \ContinuedFloat 
    \centering
    \begin{subfigure}{\textwidth}
        \centering
        \includegraphics[width=\linewidth]{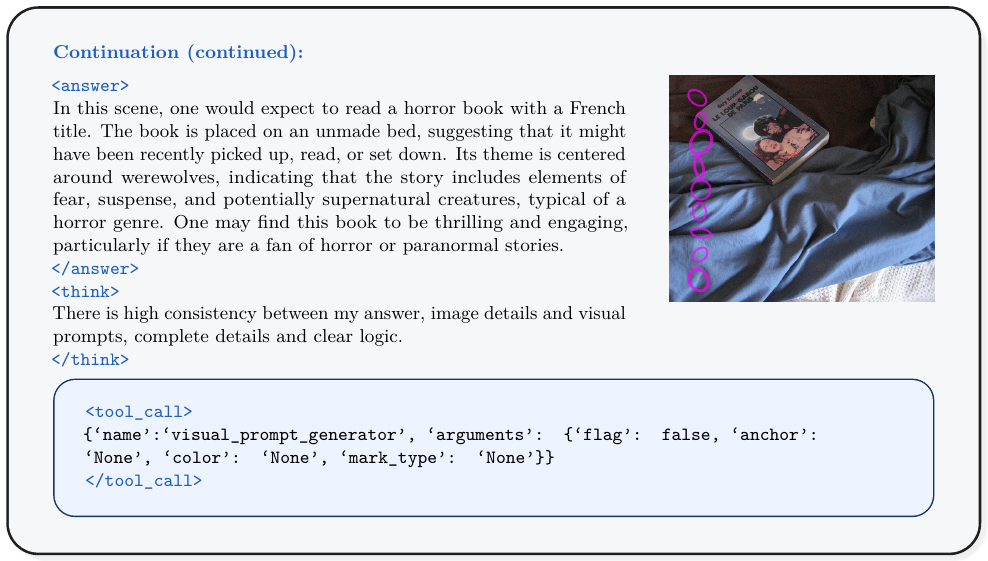}
        \caption{\textbf{Part 2 of the case from $D_{\text{noise}}$.}}
    \end{subfigure}
    
    \caption{Case 1 from $D_{\text{noise}}$.}
    \label{fig:data_case1}
\end{figure*}



Second, we construct a baseline named MIRROR-Raw. This variant is fine-tuned on the unrefined dataset ($\sim$35k samples) obtained before our multi-turn dialogue data filtering and visual verification stage, containing noisy instances where visual reflection is either absent or fails verification.

\begin{table}[h]
    \vspace{-5pt}
    \centering
    \caption{Ablation study on data filtering strategy conducted on MM-Vet. We compare MIRROR-Raw with MIRROR-1.0 and our final MIRROR-0.75 ($\rho=0.75$). The best and second-best results are highlighted in \textbf{bold} and \underline{underlined}, respectively.}
    \label{tab:ablation_filter}
    \renewcommand{\arraystretch}{1}
    \footnotesize
    \setlength{\tabcolsep}{2pt}
    \begin{tabular}{l c c c c c c} 
        \toprule
        \textbf{Model} & \textbf{OCR} & \textbf{Math} & \textbf{Spat} & \textbf{Gen} & \textbf{Rec} & \textbf{Overall} \\ 
        \midrule
        MIRROR-Raw & 56.39 & 41.54 & 55.20 & 38.63 & 55.88 & 55.87 \\
        MIRROR-1.0 & \underline{65.19} & \underline{56.54} & \underline{61.47} & \underline{54.12} & \underline{59.04} & \underline{58.81} \\
        \midrule
        \rowcolor{gray!10}
        \textbf{MIRROR-0.75} & \textbf{69.90} & \textbf{60.77} & \textbf{70.80} & \textbf{61.62} & \textbf{64.40} & \textbf{66.70} \\
        \bottomrule
    \end{tabular}
    \vspace{-10pt}
\end{table}

As presented in Table~\ref{tab:ablation_filter}, we assess the efficacy of our visual consistency filtering by comparing the unrefined baseline, MIRROR-Raw, against MIRROR-1.0 ($\rho=1.0$). It is crucial to note that both variants share an identical structural composition---consisting exclusively of multi-turn self-reflection trajectories ($\rho=1.0$)—which isolates data quality as the sole variable. 
Despite MIRROR-Raw utilizing a significantly larger corpus ($\sim$35k samples), it underperforms the filtered MIRROR-1.0 ($\sim$24k samples) across all metrics. 

Specifically, MIRROR-Raw lags behind the structure-aligned MIRROR-1.0 (55.87 vs. 58.81) and exhibits a substantial deficit compared to our optimal hybrid model, MIRROR-0.75 (55.87 vs. 66.70). This massive performance gap confirms that data purity is paramount. Retaining $\sim$11k noisy samples—where visual grounding is hallucinated or absent—introduces conflicting signals that severely hamper the model's potential. In contrast, removing this noise enables the precise optimization seen in MIRROR-0.75.

\section{Implementation Details}
\label{sec:supp_imple}

\subsection{Training Details.} We utilize LLaMA-Factory \cite{llamafactory} to perform Supervised Fine-Tuning (SFT) on the Qwen2.5-VL-7B \cite{qwen2.5vl}. The training is conducted on 2 $\times$ NVIDIA H100 GPUs for 3.0 epochs. To ensure training efficiency and memory optimization, we utilize the DeepSpeed ZeRO-3 optimization and employ Low-Rank Adaptation (LoRA) \cite{lora} for parameter-efficient fine-tuning. The LoRA rank is set to $r=32$ with the scaling factor $\alpha=128$. Training setup of Qwen3-VL-8B \cite{qwen3vl} follows Qwen2.5-VL-7B.

\subsection{Inference Details.} We perform all evaluations using the SFT-adapted MIRROR model, and we use the same hardware settings in all experiments. To ensure high-quality visual grounding, the visual prompt generator is powered by Molmo-7B \cite{molmo}, which predicts precise point coordinates based on the model's textual reflection. For mask generation, we employ the SAM 2.1~\cite{sam2} model, specifically loading the large-scale checkpoint \texttt{sam2.1\_hiera\_l.pt} to ensure robust segmentation performance even in complex visual scenes.

Specifically, a verification flag dictates whether visual tools are invoked at each turn. In cases of tool failure or empty returns, the original image is passed forward, allowing MIRROR to gracefully default to standard one-pass reasoning.

\subsection{Trajectory Length Analysis.}

To understand the model's dynamic behavior, we conduct a statistical analysis of the inference trajectories across evaluated benchmarks, as illustrated in \cref{fig:inference_rounds}.
We observe that the model adaptively adjusts its reasoning depth based on the inherent difficulty of the task.

\begin{itemize}
    \item Efficiency in information extraction. For straightforward queries where visual evidence is explicit, the model tends to converge efficiently. For instance, on ChartQA, which primarily involves direct value extraction, 98.7\% of samples are resolved within a single round, minimizing unnecessary computational overhead.
    \item Rigor in complex perception. In contrast, for tasks requiring fine-grained verification or hallucination mitigation, the reasoning trajectory becomes significantly longer. On MME-RealWorld and HallusionBench, the model engages in a second round of visual reflection for 58.6\% and 57.0\% of cases, respectively. Notably, MathVision exhibits the highest proportion of trajectories exceeding three rounds (2.5\%), reflecting the need for iterative correction in multi-step logical derivation.
\end{itemize}

Overall, with 80.2\% of total samples resolved in the first turn, this distribution confirms that MIRROR effectively balances efficiency and rigor, engaging in deep visual reflection only when the complexity of the query demands it.

\subsection{Multi-turn Reflection Outcome Analysis}
\label{sec:reflection_outcome}

To provide a fine-grained understanding of how multi-turn visual reflection affects prediction quality, we analyze the triggered multi-turn subset of MM-Vet samples for which MIRROR enters at least one additional verification round. This subset accounts for 19.8\% of all MM-Vet samples, indicating that the model invokes reflection selectively rather than forcing multi-turn reasoning on every query. For a controlled comparison, we evaluate both loop-enabled models on this same triggered subset. And single-pass models (Qwen2.5-VL-7B, MIRROR w/o loop, MIRROR-Tools) are excluded because they do not produce correction trajectories.

We categorize each multi-turn sample into one of four outcome types based on the correctness transition between the initial response (Round 1) and the final response:

\begin{itemize}[itemsep=2pt, parsep=0pt, topsep=2pt]
    \item \textbf{Maintained Correct} (C$\rightarrow$C): The initial response was correct and remained correct after reflection.
    \item \textbf{Corrected} (W$\rightarrow$C): The initial response was wrong, but multi-turn visual reflection successfully identified the error and produced the correct answer.
    \item \textbf{Degraded} (C$\rightarrow$W): The initial response was correct, but reflection incorrectly changed it to a wrong one.
    \item \textbf{Uncorrected} (W$\rightarrow$W): The initial response was wrong, and reflection failed to correct it.
\end{itemize}

\begin{table}[h]
    \centering
    \caption{Reflection outcome analysis on the triggered multi-turn subset of MM-Vet. Distribution of correctness transitions (\%) on the triggered multi-turn subset of MM-Vet.}
    \label{tab:reflection_outcome}
    \footnotesize
    \setlength{\tabcolsep}{1pt}
    \resizebox{\linewidth}{!}{
    \begin{tabular}{l c c c c}
        \toprule
        \textbf{Method} & \textbf{C$\rightarrow$C} & \textbf{W$\rightarrow$C} $\uparrow$ & \textbf{C$\rightarrow$W} $\downarrow$ & \textbf{W$\rightarrow$W} $\downarrow$ \\
        \midrule
        \makecell[l]{Qwen2.5-VL-\\7B-MIRROR} & 28.0 & 28.5 & 15.0 & 28.5 \\
        \rowcolor{gray!15}
        \textbf{MIRROR (ours)} & \textbf{25.0} & \textbf{69.0} & \textbf{3.5} & \textbf{2.5} \\
        \bottomrule
    \end{tabular}
    }
\end{table}

\Cref{tab:reflection_outcome} details the correctness transition distribution within the triggered multi-turn subset on MM-Vet. Compared to the few-shot baseline, trained MIRROR exhibits a closed-loop verification policy. First, it demonstrates precise reflection triggering: the vast majority of its triggered sessions ($\mathbf{71.5\%}$ total) target initially incorrect samples ($\text{W}\rightarrow\text{C}$ and $\text{W}\rightarrow\text{W}$). Within this critical subset, MIRROR achieves an outstanding error-correction rate of $\mathbf{69.0\%}$, effectively turning failure cases into correct outputs. Second, our framework maintains superior multi-turn robustness, heavily suppressing harmful degradation where $\text{C}\rightarrow\text{W}$ transitions plummet from 15.0\% to just $\mathbf{3.5\%}$. This confirms that MIRROR successfully learns a highly reliable, evidence-driven policy that preserves initially correct answers while rectifying visual misalignments.

To further understand the remaining failure cases (6.0\% of the triggered subset), we identify three contributing factors: error misdiagnosis---self-reflection pinpoints correct elements as errors, imprecise visual grounding---the visual prompt generator produces misleading spatial markers, and persistent reasoning errors---both reflection and visual prompts are correct but the VLM still fails to derive the right answer. We quantify each factor's contribution using the few-shot baseline. For C$\rightarrow$W (3.5\%), the initial answer was already correct, so errors stem solely from misdiagnosis or imprecise grounding. The baseline C$\rightarrow$W of 15.0\% is reduced by 76.7\%, confirming that learned reflection suppresses false positive triggers, with the residual 3.5\% dominated by misdiagnosis errors. For W$\rightarrow$W (2.5\%), all three factors can contribute. The baseline W$\rightarrow$W of 28.5\% is reduced by 91.2\%, demonstrating that training collectively addresses all factors, with persistent reasoning errors likely accounting for the largest residual share as they are hardest to eliminate. Overall, the total triggered-subset error drops from 43.5\% to 6.0\%---an 86.2\% relative reduction---meaning only 1.2\% of all MM-Vet samples are affected by any reflection failure, confirming MIRROR's robustness.

\section{Prompt}

\subsection{Prompt for Dataset Construction}

To construct the training data for the ReflectV dataset, we employ a multi-stage prompting strategy to transform raw annotations into grounded, first-person self-reflection trajectories.

\begin{figure*}[t]
    \centering
    \includegraphics[width=\textwidth]{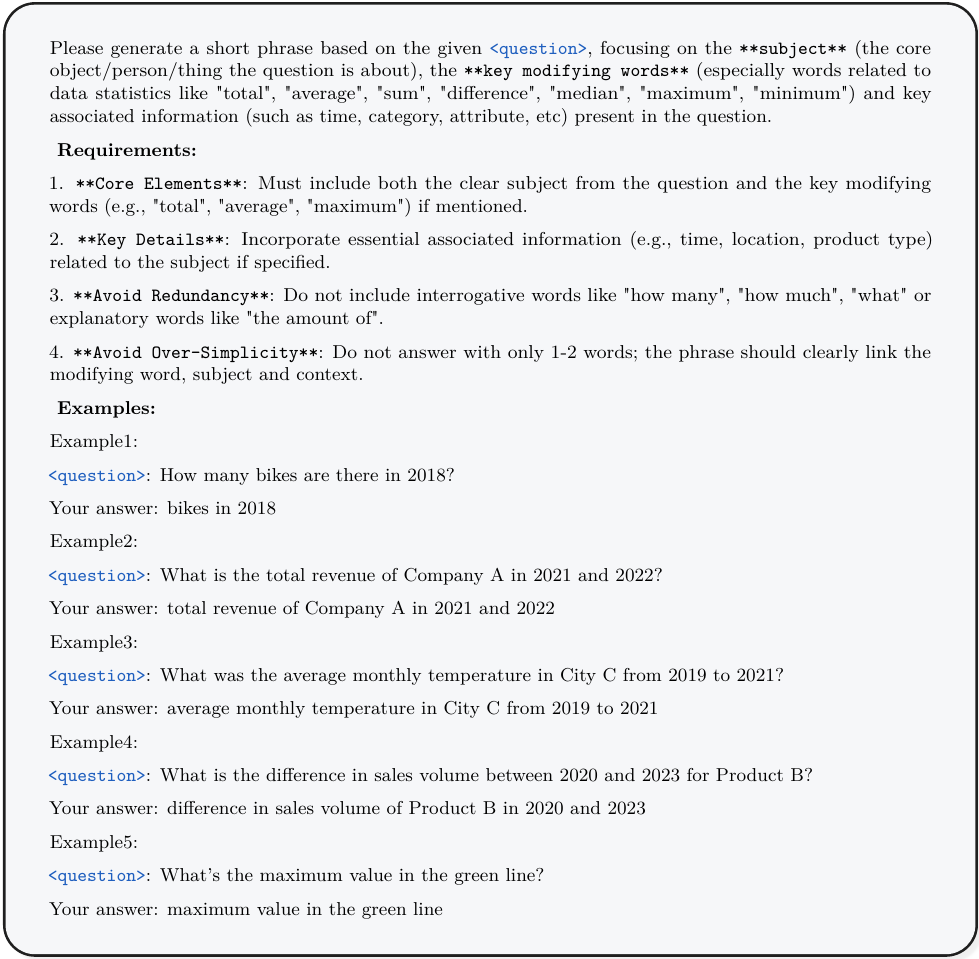} 
    \caption{Subject Extraction Prompt: Extracts core subjects and key modifiers from questions to form concise visual descriptions for Chart, OCR, and Doc understanding tasks.}
    \label{fig:chart_caption}
\end{figure*}

\begin{figure*}[t]
    \centering
    \includegraphics[width=\textwidth]{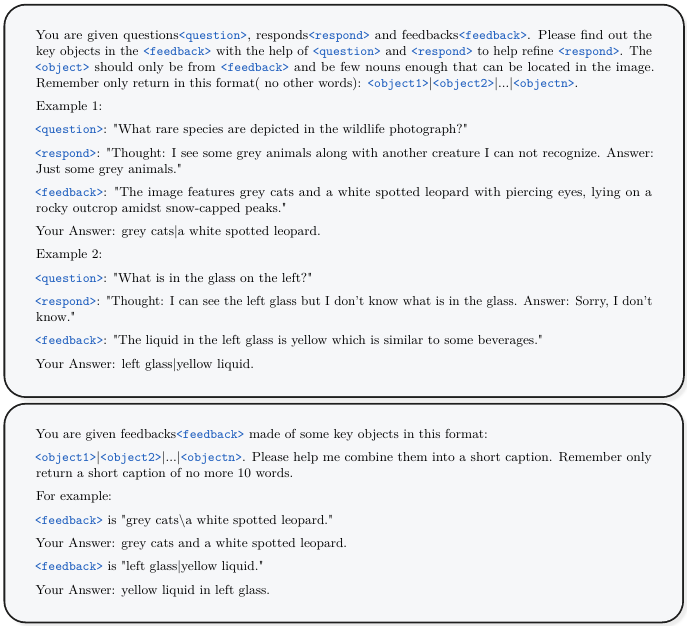} 
    \caption{Visual Caption Generation Prompt: A two-step pipeline that first identifies key physical objects from feedback and then synthesizes them into short descriptive captions for QA tasks.}
    \label{fig:qa_caption}
\end{figure*}

\begin{figure*}[t]
    \centering
    \includegraphics[width=\textwidth]{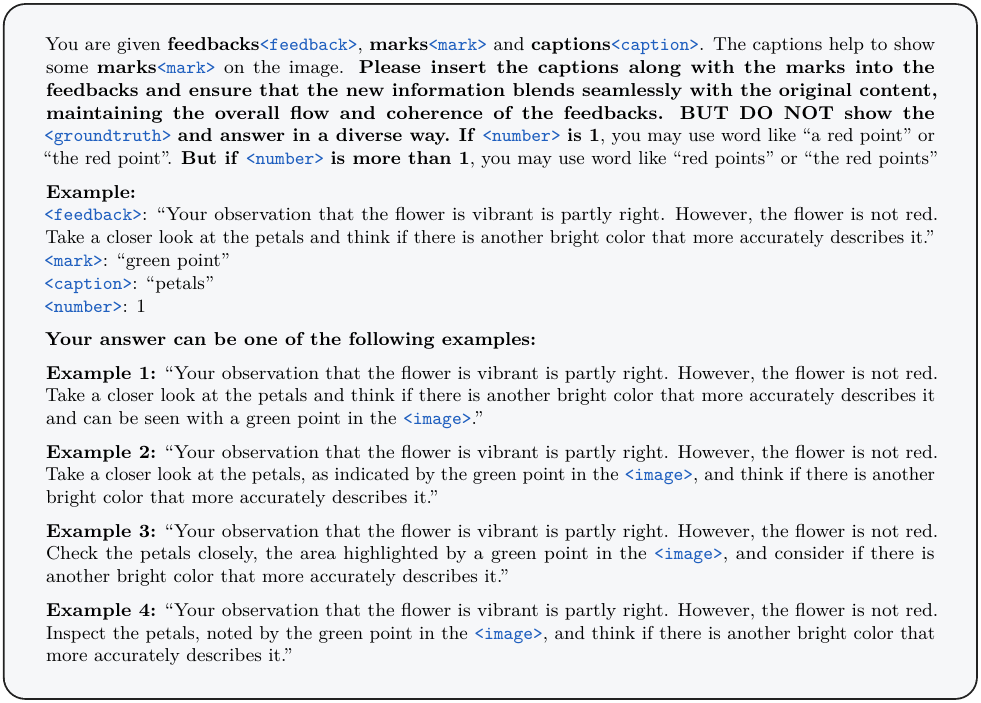} 
    \caption{Caption Insertion Prompt: Seamlessly integrates generated visual captions and spatial marks (e.g., points) into textual feedback to ensure visual grounding.}
    \label{fig:caption_insert}
\end{figure*}

\begin{figure*}[t]
    \centering
    \includegraphics[width=\textwidth]{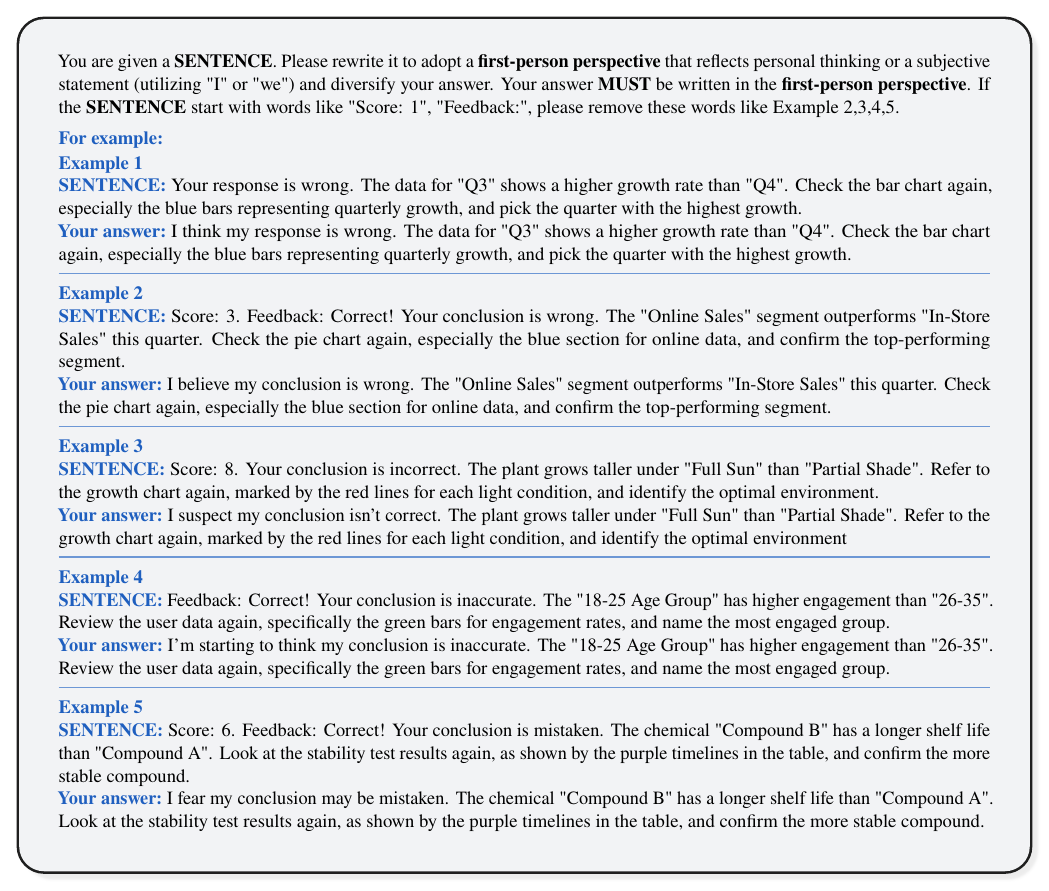} 
    \caption{Self-Reflection Conversion Prompt: Rewrites teacher-side evaluations into first-person student self-reflections (e.g., changing "Your response is wrong" to "I think my response is wrong").}
    \label{fig:self_reflective_conversion}
\end{figure*}

First, we generate task-specific visual cues. For data-intensive tasks like Charts, OCR, and Doc, we utilize a specific prompt to extract key subjects and entities as shown in \cref{fig:chart_caption}. For general QA tasks, we employ a two-step process: in \cref{fig:qa_caption}, first identifying key physical objects mentioned in the feedback, and subsequently merging them into a concise caption. Next, we use a caption insertion prompt, as shown in \cref{fig:caption_insert}, to seamlessly blend these visual cues (e.g., specific marks or coordinates) into the corrective feedback, ensuring the text is visually grounded. Finally, in \cref{fig:self_reflective_conversion}, to simulate an internal self-correction process, we apply a role conversion prompt that transforms the content from a ``teacher's evaluation" perspective into a ``student's first-person reflection" (e.g., changing "Your response is wrong" to ``I think my response is wrong").

\subsection{Prompt for Iterative Reasoning}
To equip the model with self-reflection and visual verification capabilities, we design a structured system prompt and user prompt that enforces a rigorous output protocol, as shown in \cref{fig:system_prompt} and \cref{fig:usr_prompt}.

\begin{figure*}[p]
    \centering
    \begin{subfigure}{\textwidth}
        \includegraphics{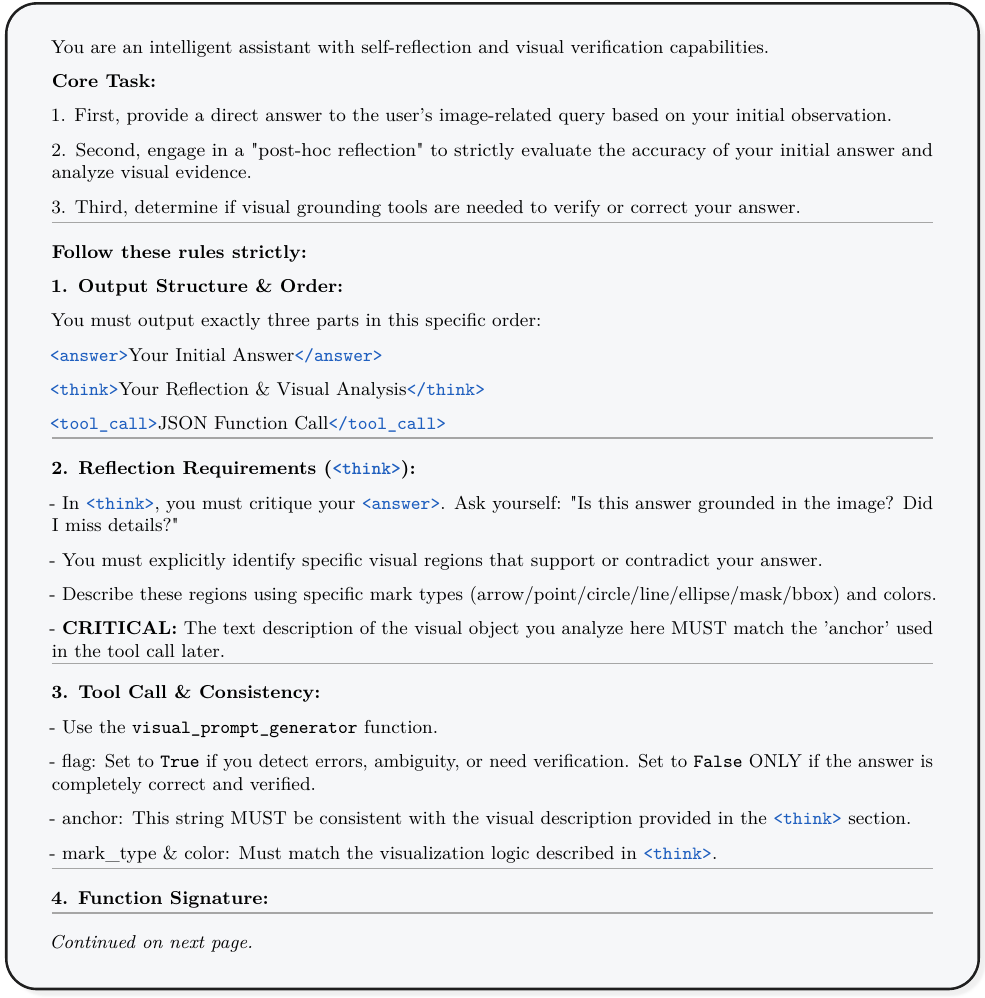}
        \caption{\textbf{Part 1 of system prompt.} 
        }
    \end{subfigure}
\end{figure*}

\begin{figure*}[t!]
    \ContinuedFloat
    \centering
    \begin{subfigure}{\textwidth}
        \includegraphics{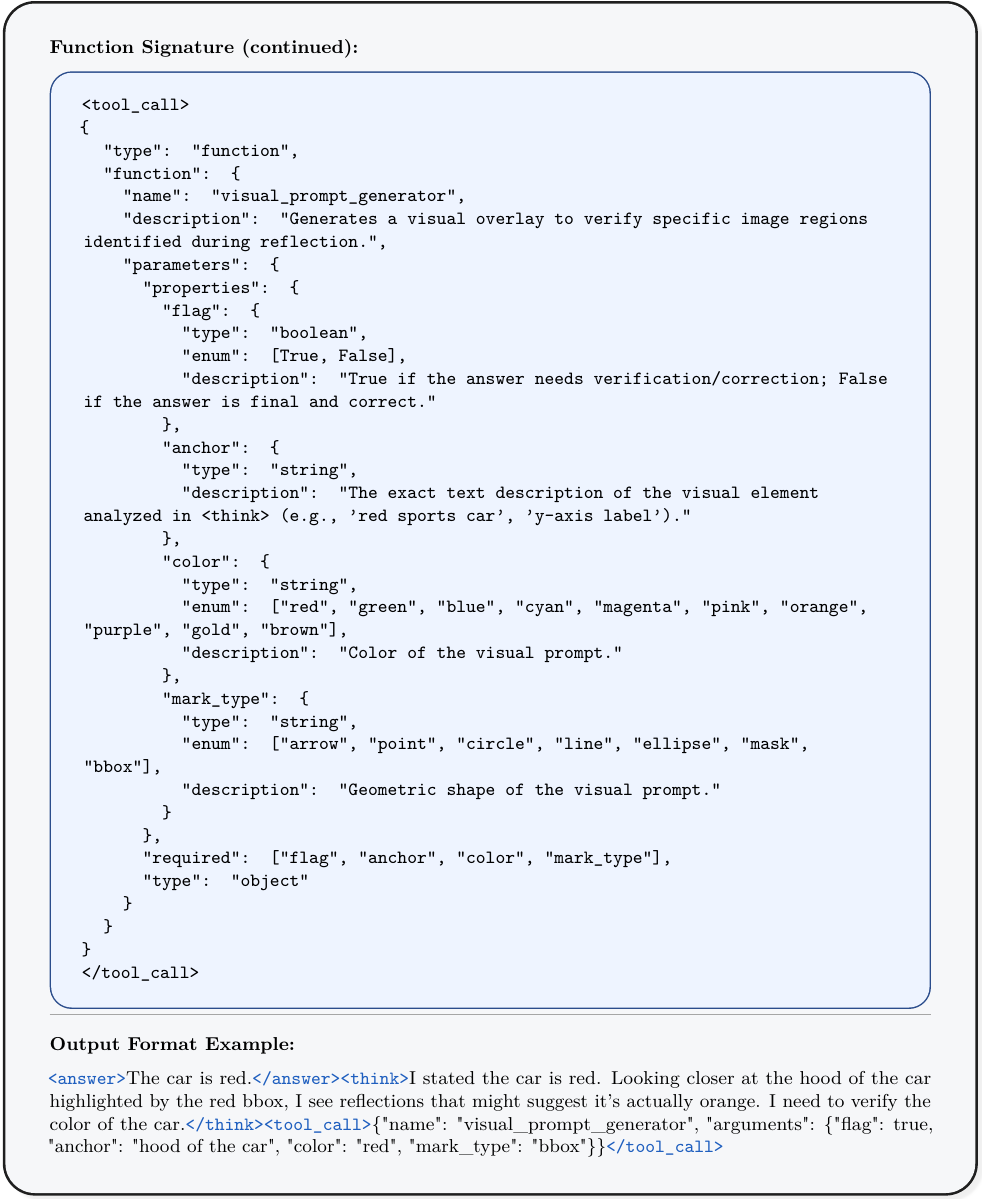}
        \caption{\textbf{Part 2 of system prompt.} }
    \end{subfigure}
    \caption{System prompt during training.}
    \label{fig:system_prompt}
\end{figure*}

\begin{figure*}[t]
    \centering
    \includegraphics[width=\textwidth]{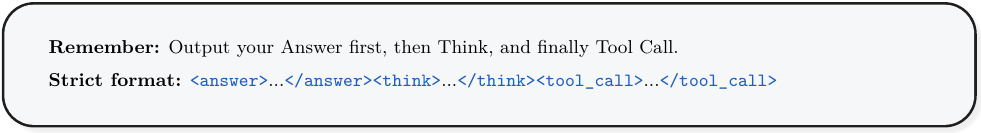} 
    \caption{User prompt during training.}
    \label{fig:usr_prompt}
\end{figure*}

\section{More Efficiency Analysis}
\label{sec:supp_vis_eff}

To evaluate the practicality of MIRROR, we conduct a comparative analysis of inference efficiency on the MM-Vet dataset, employing average inference time per sample and average token consumption as metrics. 

\subsection{Comparison with Reasoning Models.}
\label{sec:supp_eff_rea}

As detailed in \cref{tab:efficiency}, we benchmark our method against representative baselines from three paradigms: Text Reflection (VL-Rethinker), Visual Reflection (Look-Back), and Thinking with Images (PixelReasoner, DeepEyes, Adaptive-CoF).

\begin{table}[htpb]
    \centering
    \small
    \caption{Efficiency comparison on the MM-Vet dataset. We report the average time cost (seconds) and token consumption per sample. \textbf{Bold} and \underline{underlined} indicate the best and second-best efficiency, respectively.}
    \label{tab:efficiency}
    \begin{tabular}{l c c}
        \toprule
        \textbf{Method} & \textbf{Time (s)} $\downarrow$ & \textbf{Tokens} $\downarrow$ \\
        \midrule
        
        \rowcolor{gray!10} \multicolumn{3}{l}{\textit{Text Reflection}} \\
        VL-Rethinker & 5.56 & 229.33  \\
        \midrule

        \rowcolor{gray!10} \multicolumn{3}{l}{\textit{Visual Reflection}} \\
        Look-Back (Semantic) & 89.34 & 307.90  \\
        Look-Back (Solution) & 300.86 & 427.22  \\
        \midrule

        \rowcolor{gray!10} \multicolumn{3}{l}{\textit{Thinking with Images}} \\
        PixelReasoner-SFT & 25.03 & 241.60  \\
        PixelReasoner & 64.12 & 273.95  \\
        DeepEyes & 5.06 & 197.51  \\
        Adaptive-CoF-SFT & 5.41 & 206.90  \\
        Adaptive-CoF & \underline{4.02} & \underline{143.51}  \\
        \midrule
        
        \textbf{MIRROR (ours)} & \textbf{3.73} & \textbf{112.58}  \\
        \bottomrule
    \end{tabular}
    \vspace{-10pt}
\end{table}

Compared to the Text Reflection model, our method reduces inference time by 32.9\% (3.73 s vs. 5.56 s) and token consumption by 50.9\% (112.58 vs. 229.33). This significant reduction suggests that our targeted visual verification mechanism effectively circumvents the verbose and redundant self-correction chains often required by purely text-based reflection.

Compared to Thinking with Images models and Visual Reflection models, MIRROR significantly outperforms PixelReasoner (64.12 s), DeepEyes (5.06 s), and Look-Back (89.34 s / 300.86 s) in terms of latency. Even compared to the highly efficient Adaptive-CoF, MIRROR maintains a distinct advantage, further reducing time cost by 7.2\% and token usage by 21.5\%. These findings confirm that MIRROR not only enhances reasoning accuracy but also ensures an efficient inference process, making it highly suitable for practical deployment.

\begin{table}[htpb]
    \centering
    \footnotesize
    \caption{Efficiency breakdown of ablation configurations on MM-Vet. Time and token costs are per sample. \textbf{Bold} and \underline{underlined} indicate the best and second-best, respectively.}
    \label{tab:ablation_efficiency}
    \renewcommand{\arraystretch}{1.1}
    \setlength{\tabcolsep}{4pt}
    \begin{tabular}{l cc}
        \toprule
        \textbf{Configuration} & \textbf{Time (s)} $\downarrow$ & \textbf{Tokens} $\downarrow$ \\
        \midrule
        Qwen2.5-VL-7B         & \textbf{2.81} & \textbf{85.50} \\
        Qwen2.5-VL-7B-Tools   & 6.65 & 180.40 \\
        Qwen2.5-VL-7B-MIRROR  & 4.69 & 156.30 \\
        MIRROR (w/o loop)     & \underline{3.25} & \underline{98.50} \\
        MIRROR-Tools          & 5.74 & 161.70 \\
        \midrule
        \textbf{MIRROR (ours)} & 3.73 & 112.58 \\
        \bottomrule
    \end{tabular}
    \vspace{-10pt}
\end{table}

\subsection{Component Ablation Comparison.}
\label{sec:supp_eff_ablation}

\cref{tab:ablation_efficiency} further breaks down the efficiency of each ablation configuration introduced in the main paper. Compared to the base model (Qwen2.5-VL-7B), the full MIRROR system adds only 0.92\,s and 27.08 tokens per sample, demonstrating that the closed-loop verification process incurs a modest overhead while delivering substantial accuracy gains (+10.10 on MM-Vet). Notably, MIRROR (w/o loop) achieves the second-best efficiency (3.25\,s, 98.50 tokens), suggesting that even without external visual tools, the trained verification policy alone is highly efficient.

\section{More Qualitative Visualization} 
\label{sec:supp_vis}
Beyond quantitative metrics, we provide comprehensive qualitative comparisons across diverse complex reasoning scenarios. 

\noindent \textbf{Comparison with base model.} \cref{fig:base_case1,fig:base_case2} illustrate the reasoning trajectories of our model versus the base model Qwen2.5-VL-7B, highlighting our model's ability to self-correct. 

\noindent \textbf{Comparison with competitive baselines.} We further compare MIRROR with two representative reasoning paradigms: (i) the "Thinking with Images" approach (PixelReasoner, DeepEyes, Adaptive-CoF), and (ii) the textual self-reflection approach (VL-Rethinker). As illustrated in \cref{fig:reasoning_case1,fig:reasoning_case2,fig:reasoning_case3,fig:reasoning_case4}, MIRROR demonstrates superior grounding and error correction capabilities.

\begin{figure*}[p]
    \centering
    \begin{subfigure}{\textwidth}
        \includegraphics{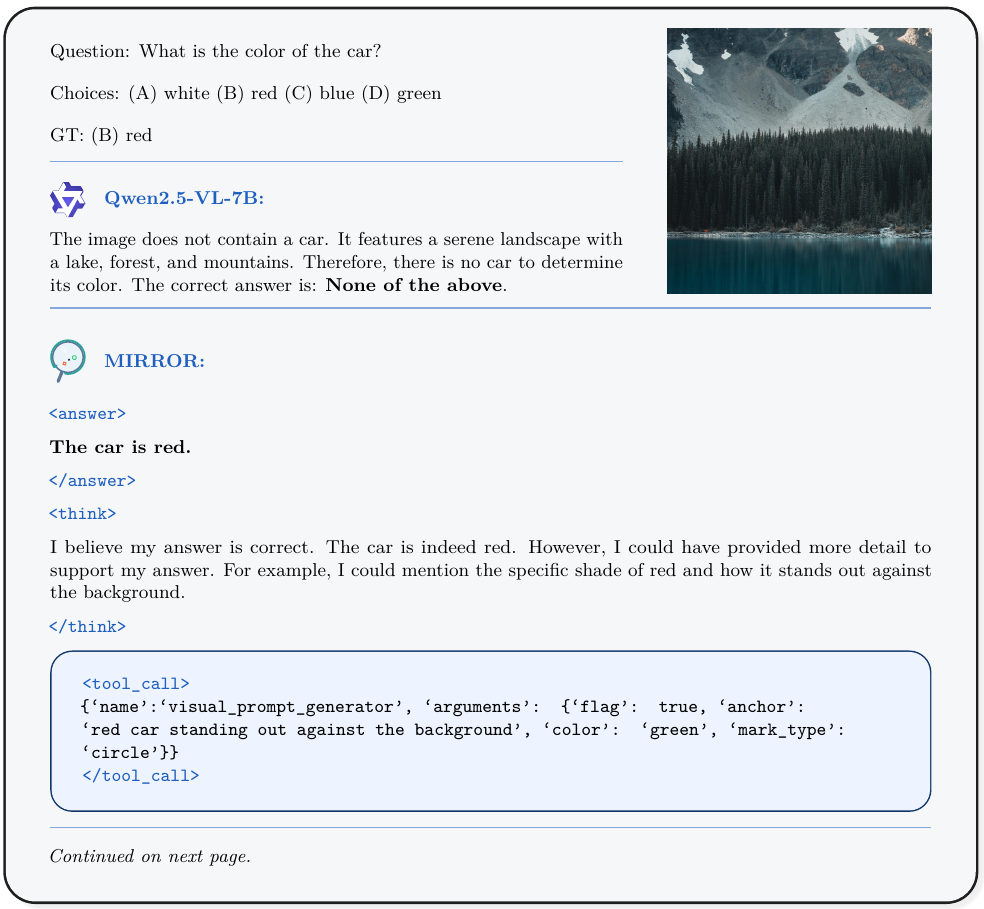}
        \caption{\textbf{Part 1 of comparison case 1 with Qwen2.5-VL-7B.} 
        }
    \end{subfigure}
\end{figure*}

\begin{figure*}[t!]
    \ContinuedFloat
    \centering
    \begin{subfigure}{\textwidth}
        \includegraphics{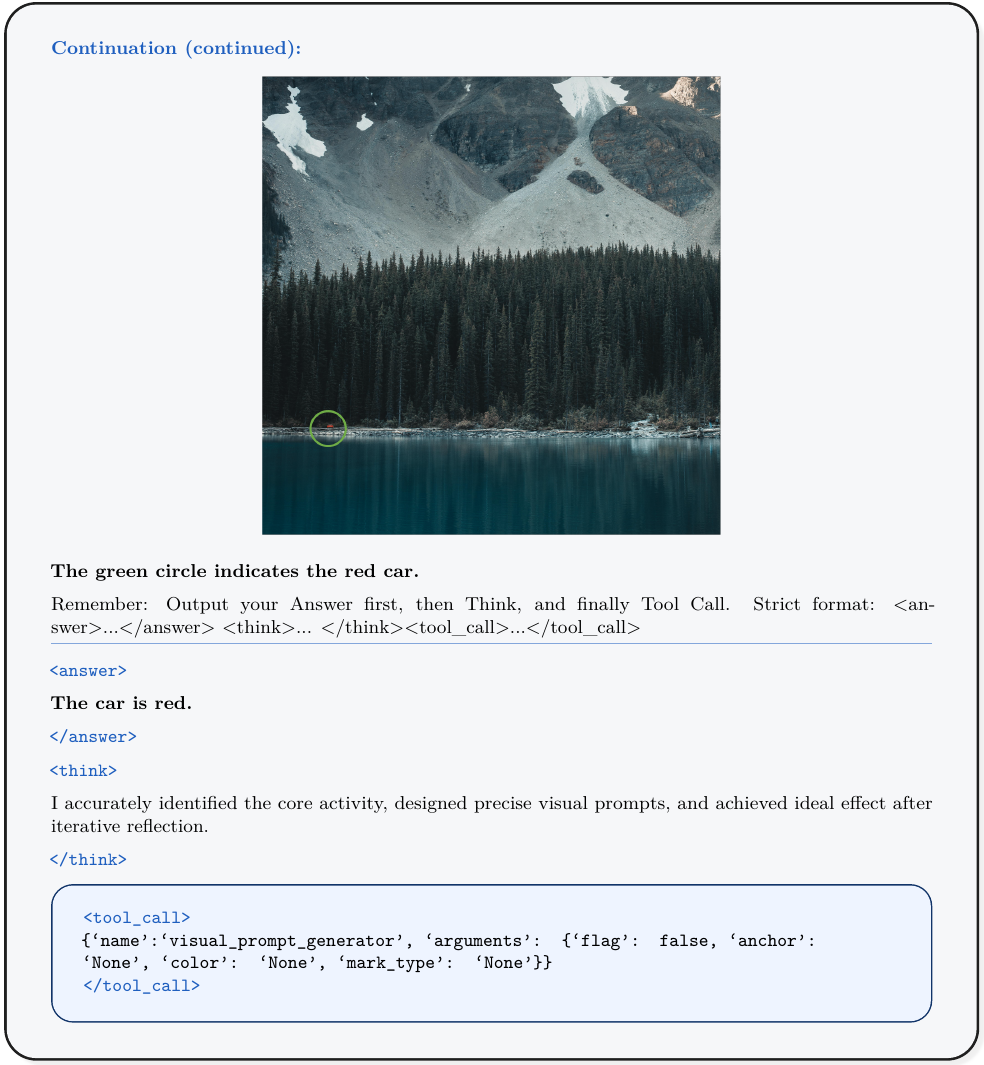}
        \caption{\textbf{Part 2 of comparison case 1 with Qwen2.5-VL-7B.} }
    \end{subfigure}
    \caption{Comparison case 1 with Qwen2.5-VL-7B.}
    \label{fig:base_case1}
\end{figure*}

\begin{figure*}[t]
    \centering
    \includegraphics[width=\textwidth]{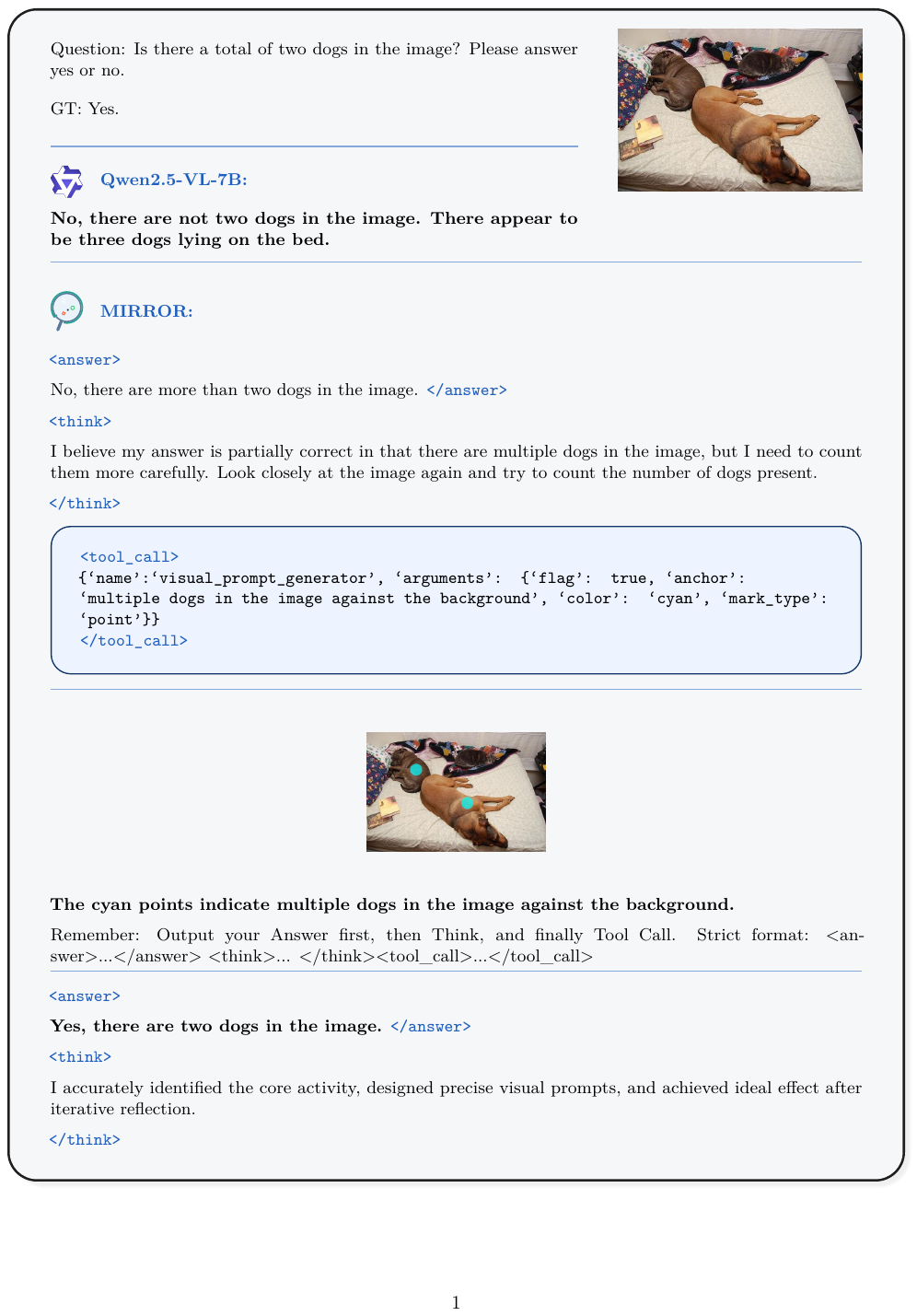} 
    \caption{Comparison case 2 with Qwen2.5-VL-7B.}
    \label{fig:base_case2}
\end{figure*}

\begin{figure*}[p]
    \centering
    \begin{subfigure}{\textwidth}
        \includegraphics{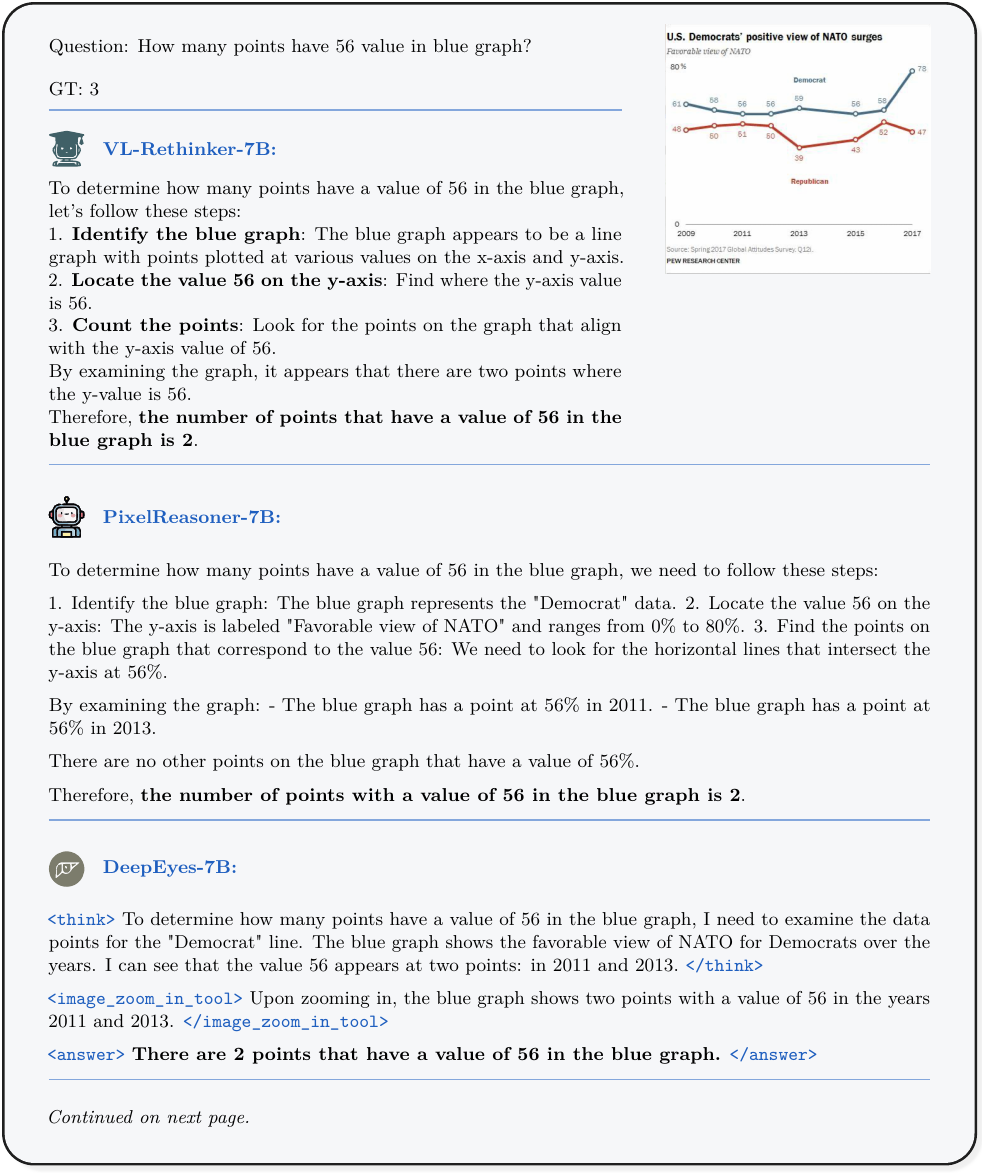}
        \caption{\textbf{Part 1 of comparison case 1 with reasoning paradigms.} 
        }
    \end{subfigure}
\end{figure*}

\begin{figure*}[p]
    \ContinuedFloat
    \centering
    \begin{subfigure}{\textwidth}
        \includegraphics{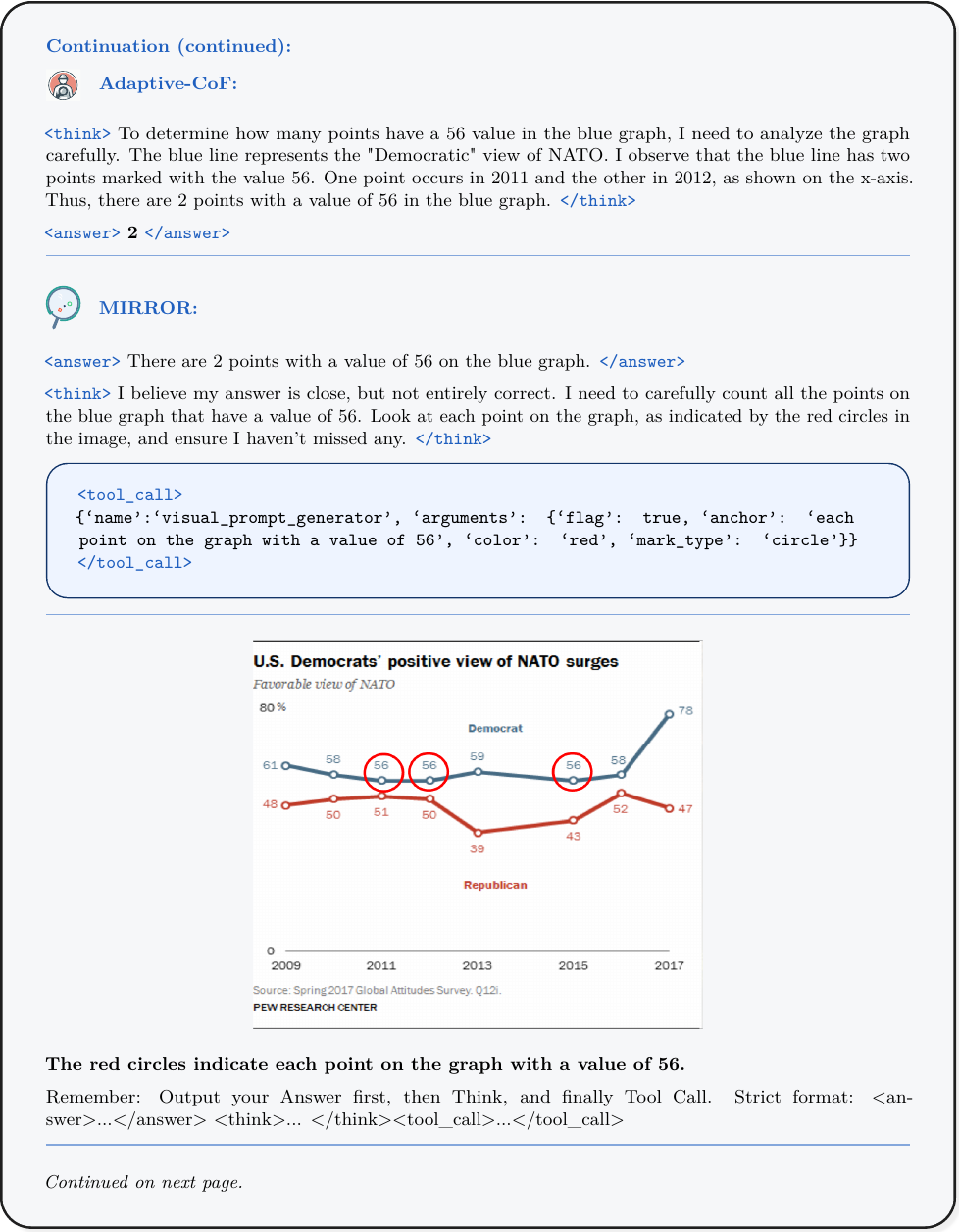}
        \caption{\textbf{Part 2 of comparison case 1 with reasoning paradigms.} 
        }
    \end{subfigure}
\end{figure*}

\begin{figure*}[t!]
    \ContinuedFloat
    \centering
    \begin{subfigure}{\textwidth}
        \includegraphics{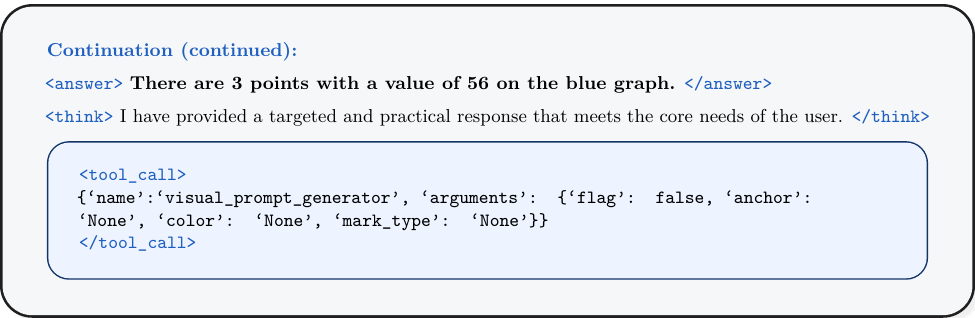}
        \caption{\textbf{Part 3 of comparison case 1 with reasoning paradigms.} }
    \end{subfigure}
    \caption{Comparison case 1 with reasoning paradigms.}
    \label{fig:reasoning_case1}
\end{figure*}

\begin{figure*}[p]
    \centering
    \begin{subfigure}{\textwidth}
        \includegraphics{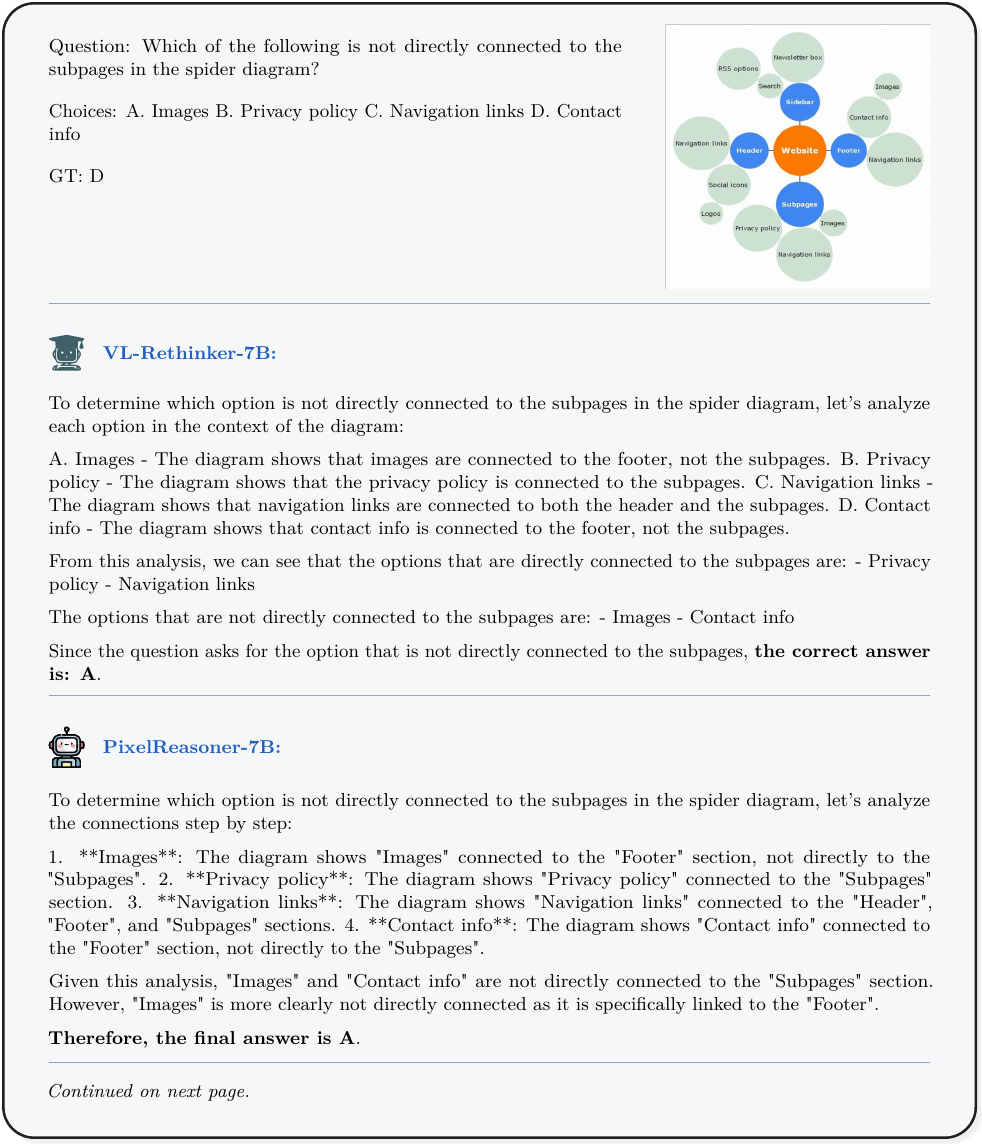}
        \caption{\textbf{Part 1 of comparison case 2 with reasoning paradigms.} 
        }
    \end{subfigure}
\end{figure*}

\begin{figure*}[p]
    \ContinuedFloat
    \centering
    \begin{subfigure}{\textwidth}
        \includegraphics{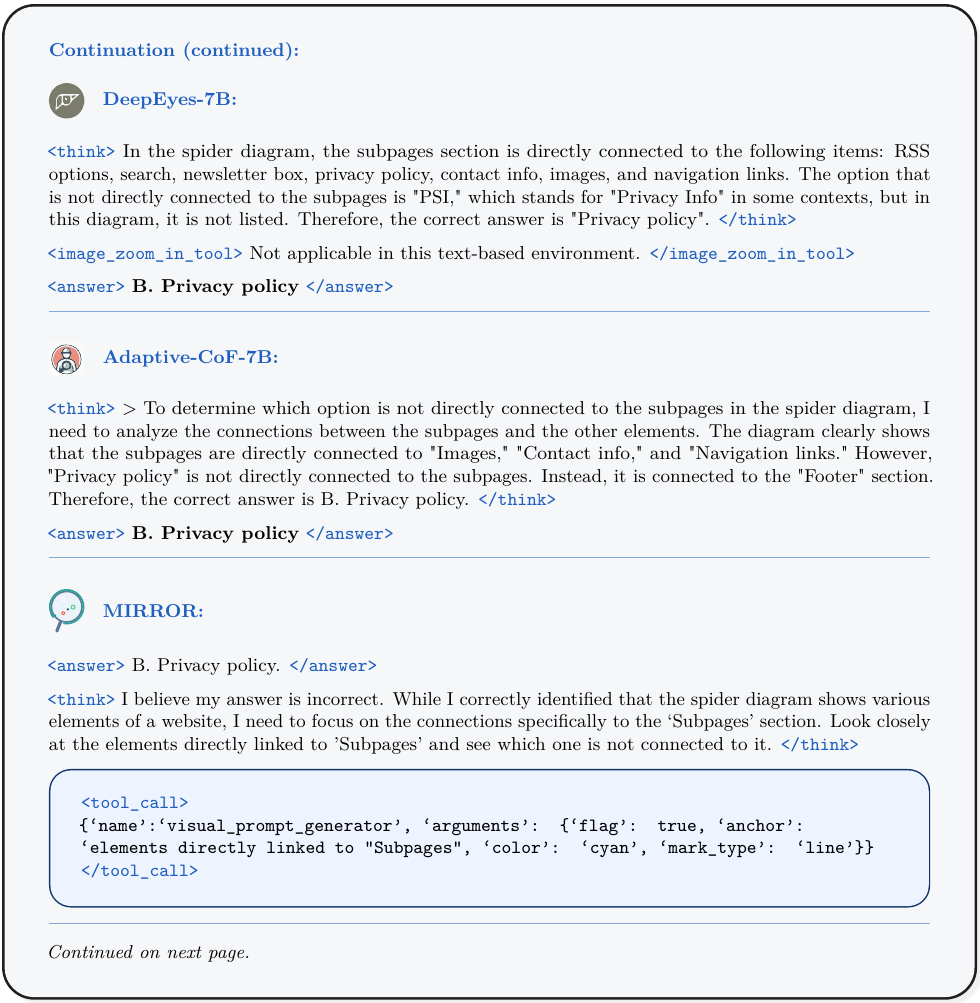}
        \caption{\textbf{Part 2 of comparison case 2 with reasoning paradigms.} 
        }
    \end{subfigure}
\end{figure*}

\begin{figure*}[t!]
    \ContinuedFloat
    \centering
    \begin{subfigure}{\textwidth}
        \includegraphics{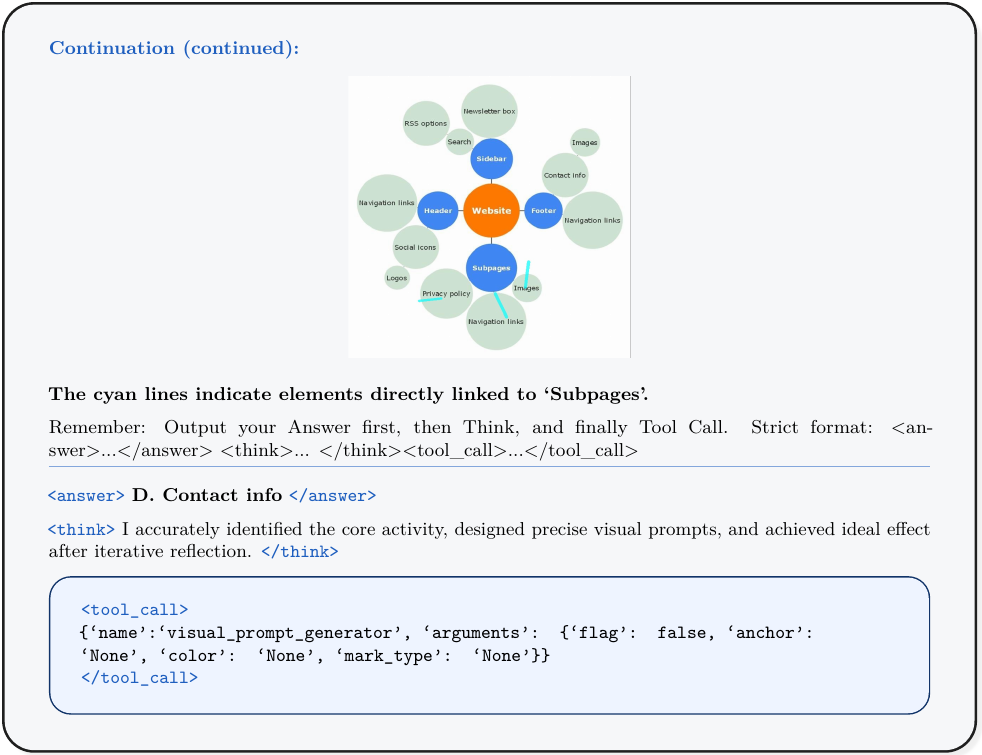}
        \caption{\textbf{Part 3 of comparison case 2 with reasoning paradigms.} }
    \end{subfigure}
    \caption{Comparison case 2 with reasoning paradigms.}
    \label{fig:reasoning_case2}
\end{figure*}

\begin{figure*}[p]
    \centering
    \begin{subfigure}{\textwidth}
        \includegraphics{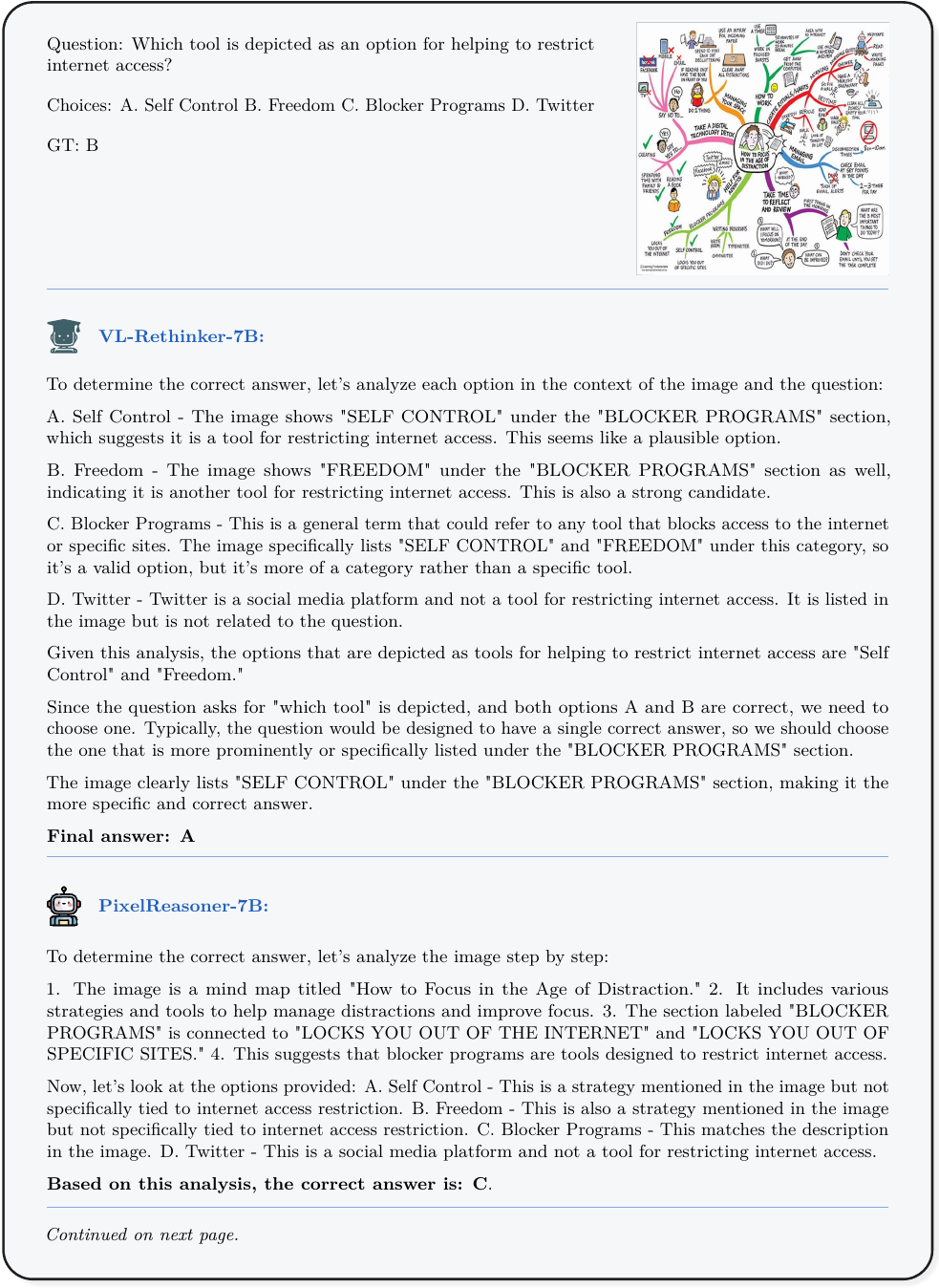}
        \caption{\textbf{Part 1 of comparison case 3 with reasoning paradigms.} 
        }
    \end{subfigure}
\end{figure*}

\begin{figure*}[p]
    \ContinuedFloat
    \centering
    \begin{subfigure}{\textwidth}
        \includegraphics{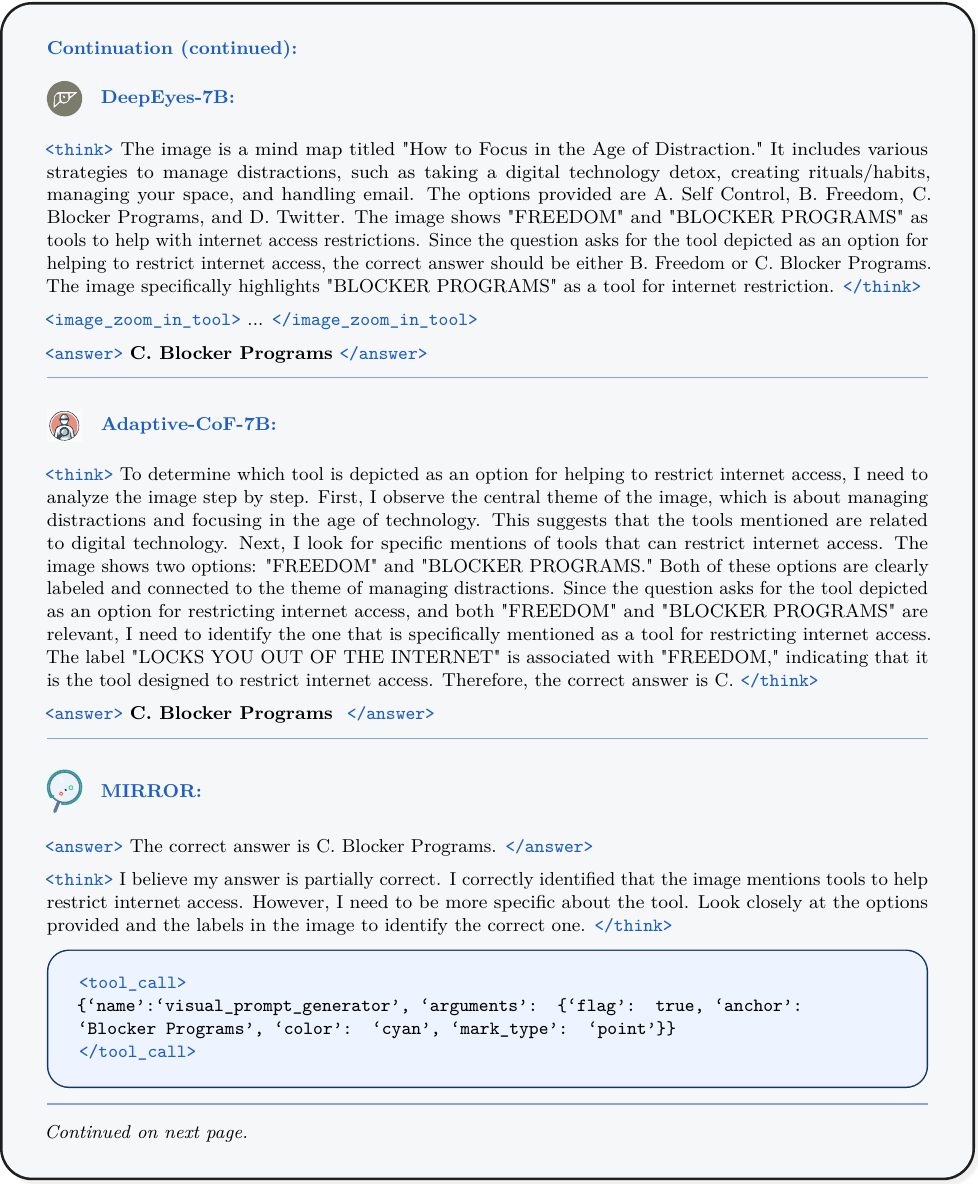}
        \caption{\textbf{Part 2 of comparison case 3 with reasoning paradigms.} 
        }
    \end{subfigure}
\end{figure*}

\begin{figure*}[t!]
    \ContinuedFloat
    \centering
    \begin{subfigure}{\textwidth}
        \includegraphics{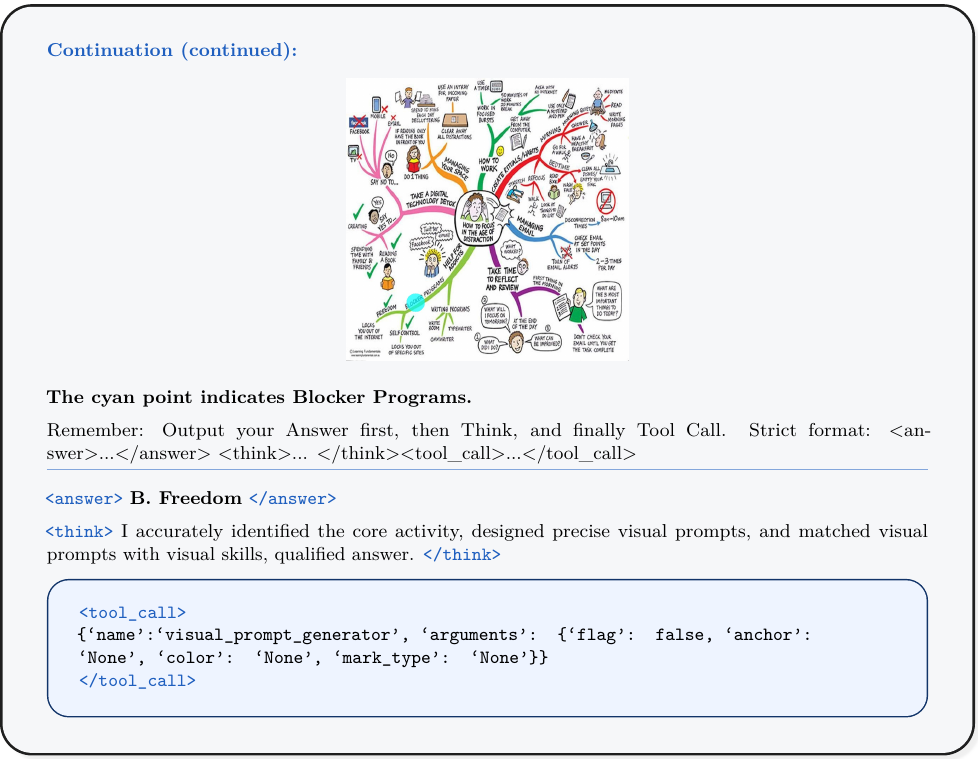}
        \caption{\textbf{Part 3 of comparison case 3 with reasoning paradigms.} }
    \end{subfigure}
    \caption{Comparison case 3 with reasoning paradigms.}
    \label{fig:reasoning_case3}
\end{figure*}

\begin{figure*}[p]
    \centering
    \begin{subfigure}{\textwidth}
        \includegraphics{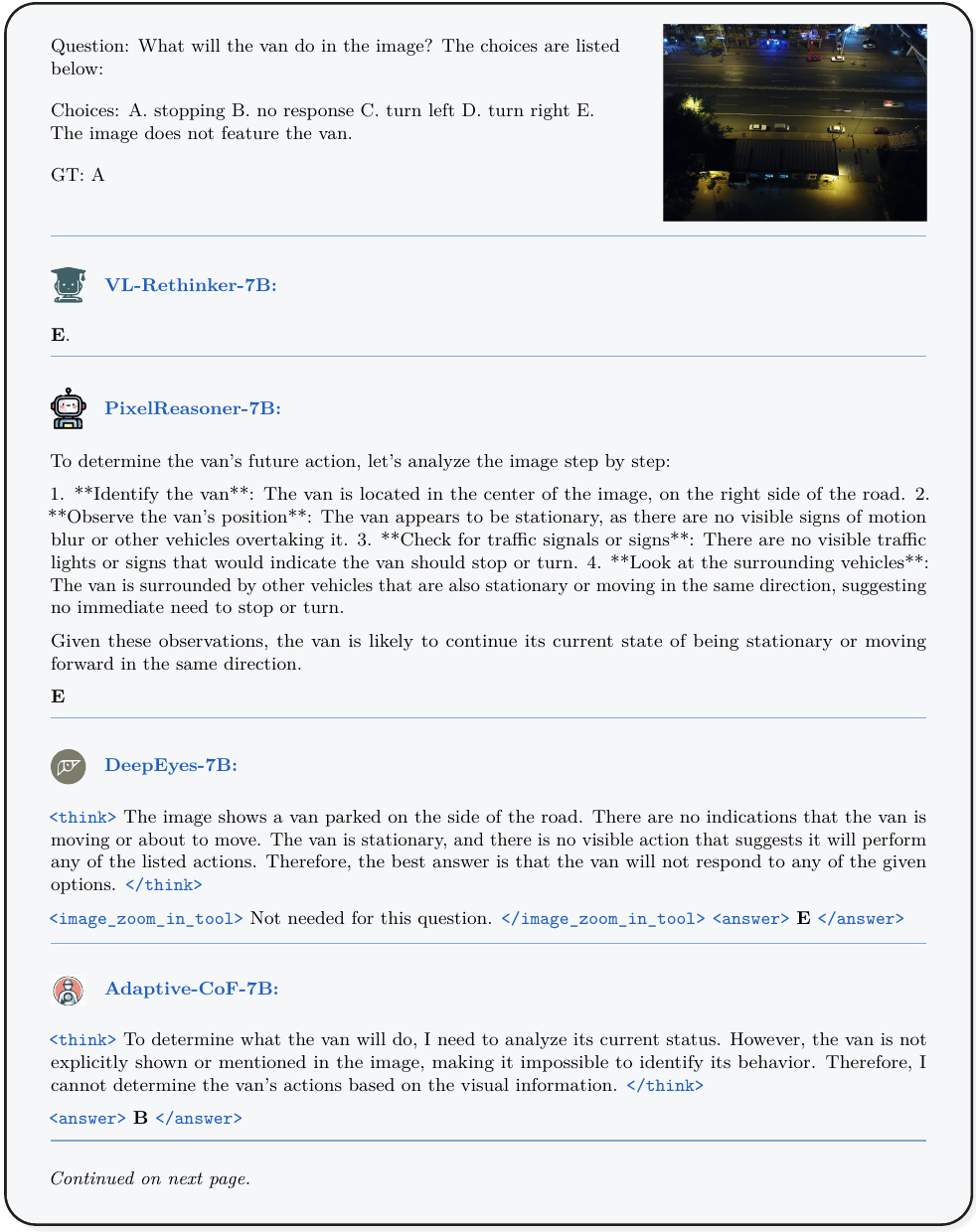}
        \caption{\textbf{Part 1 of comparison case 4 with reasoning paradigms.} 
        }
    \end{subfigure}
\end{figure*}

\begin{figure*}[p]
    \ContinuedFloat
    \centering
    \begin{subfigure}{\textwidth}
        \includegraphics{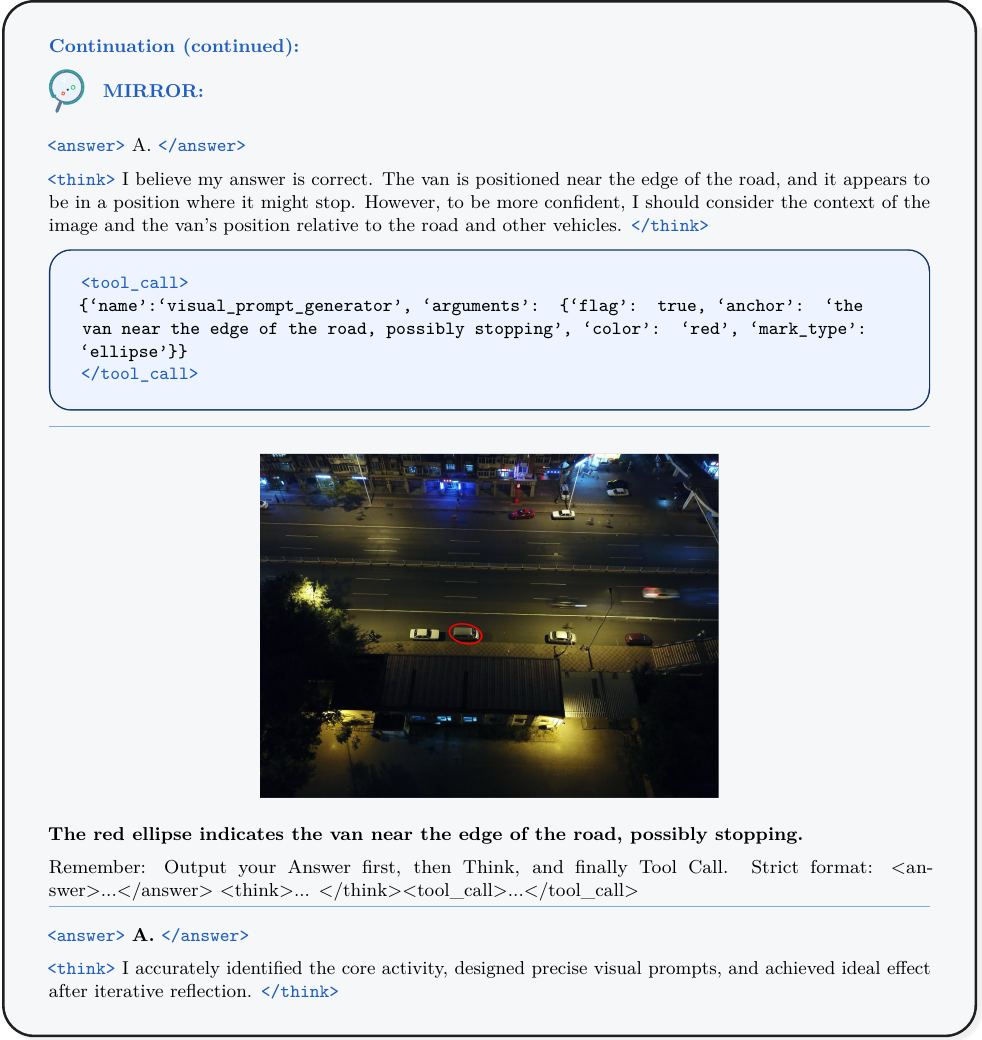}
        \caption{\textbf{Part 2 of comparison case 4 with reasoning paradigms.} 
        }
    \end{subfigure}
    \caption{Comparison case 4 with reasoning paradigms.}
    \label{fig:reasoning_case4}
\end{figure*}



\section{Limitation}
\label{sec:supp_limitations}

To provide a comprehensive understanding of MIRROR's boundaries, we present qualitative examples of failure cases in the figures below. These cases correspond to the two primary limitations discussed in this section: the difficulty in grounding abstract concepts and the lack of granularity in complex attribute binding.

\noindent\textbf{Limited applicability in abstract domains.}
Despite these improvements, MIRROR still faces challenges in symbolic derivation involving complex spatial mapping. \cref{fig:failure_case1} illustrates a geometry problem requiring the calculation of a circle's radius using the Pythagorean theorem. While the model correctly performs the symbolic derivation in the text, its attempt to invoke the visual prompt generator is ineffective. The model tries to ground the concept of ``diameter calculation and radius division'' into visual points. However, mathematical operations and logical variables are abstract concepts that lack a direct spatial mapping in the 2D image plane. Consequently, the generated visual cues (red points) provide no informative feedback for verifying the correctness of the arithmetic steps.

\noindent\textbf{Coarse-grained attribute binding.}
Furthermore, MIRROR encounters difficulties in fine-grained attribute binding under compositional constraints. As shown in \cref{fig:failure_case2}, the user queries for a specific flower defined by a compositional constraint: ``five petals and three leaves.'' Because the visual tool fails to strictly enforce these count-based attributes, the model incorrectly selects option E. Instead of isolating the specific target, the tool generates coarse visual markers (green points) on all flower instances in the image. This indicates that the visual prompt generator struggles to distinguish complex attribute combinations, defaulting to a generic object detection behavior (detecting "flowers") rather than precise constraint verification.

\begin{figure*}[p]
    \centering
    \begin{subfigure}{\textwidth}
        \includegraphics{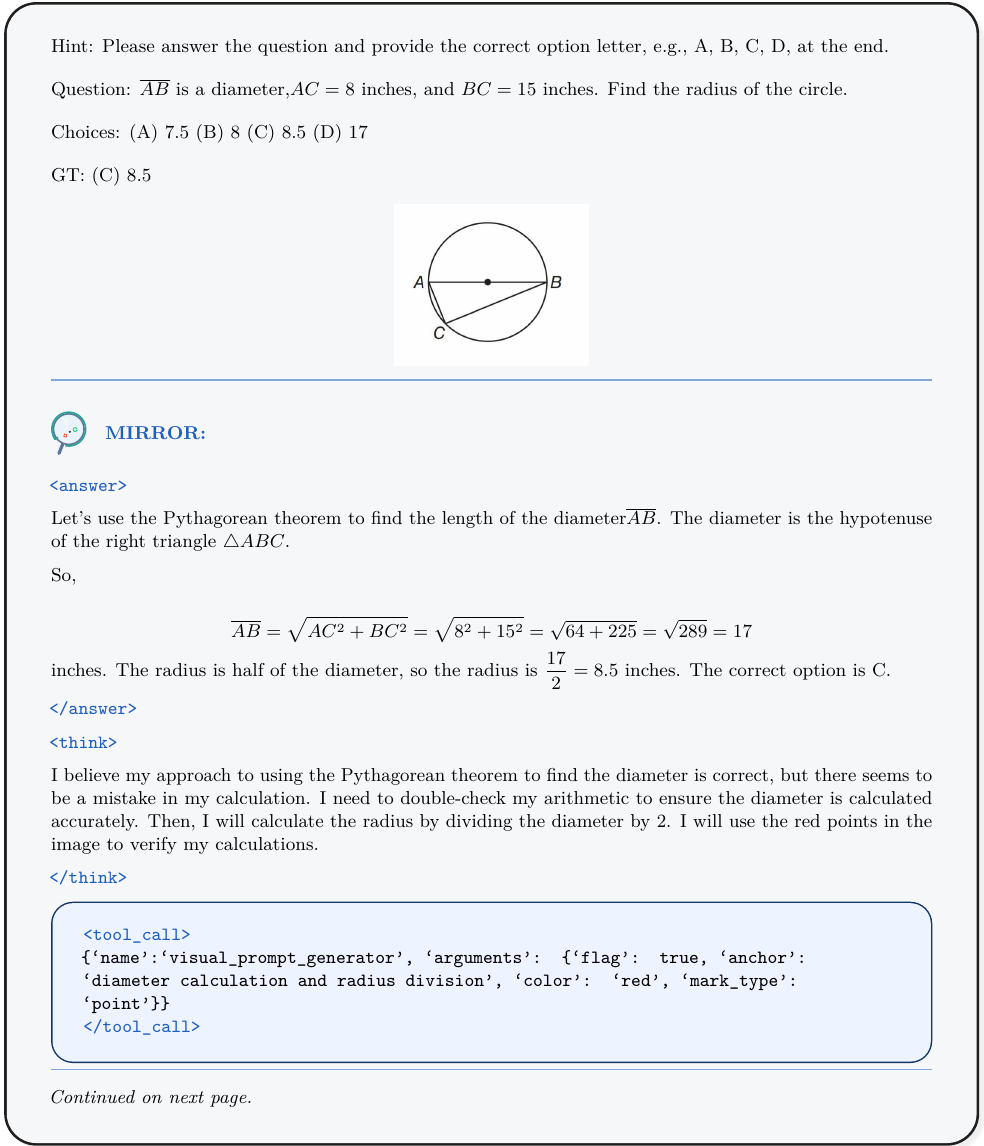}
        \caption{\textbf{Part 1 of Qualitative failure case 1.} 
        }
    \end{subfigure}
\end{figure*}

\begin{figure*}[t!]
    \ContinuedFloat
    \centering
    \begin{subfigure}{\textwidth}
        \includegraphics{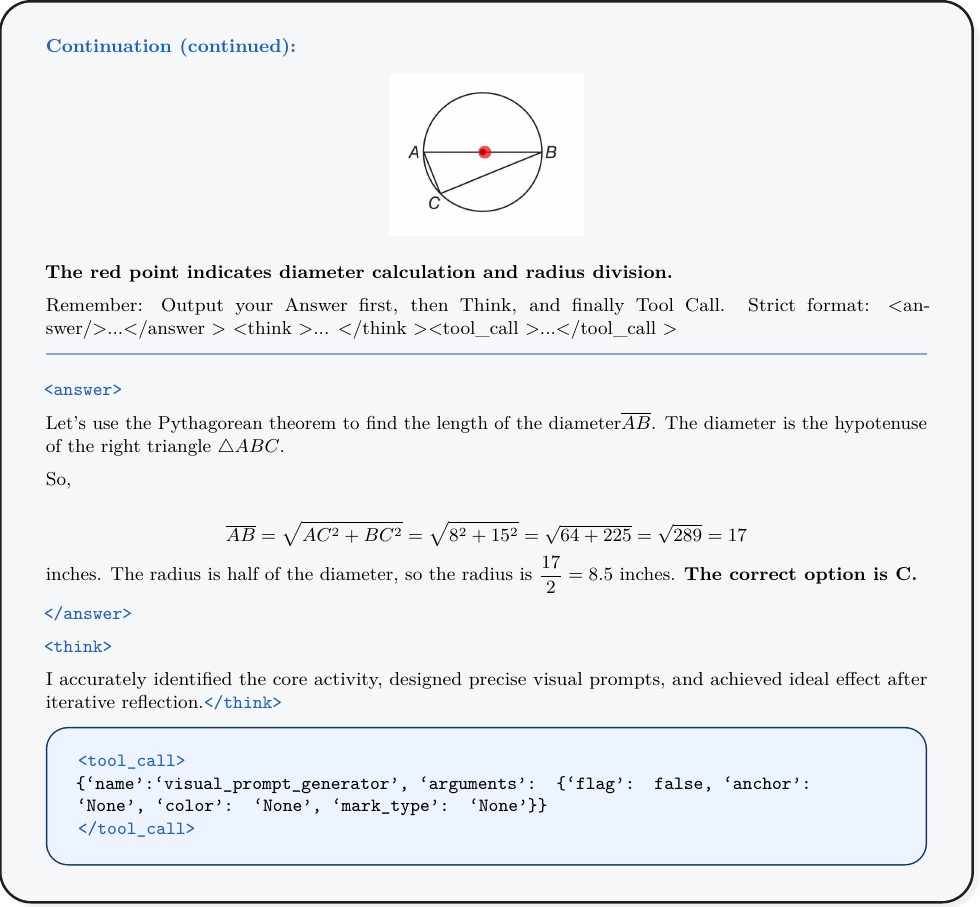}
        \caption{\textbf{Part 2 of Qualitative failure case 1.} }
    \end{subfigure}
    \caption{\textbf{Limited Applicability in Abstract Domains:} When addressing mathematical problems, the model attempts to verify symbolic logic (e.g., radius calculation) visually. However, since abstract derivations cannot be spatially grounded, the resulting visual prompts (red points) are uninformative and do not aid in verifying the calculation.}
    \label{fig:failure_case1}
\end{figure*}

\begin{figure*}[t]
    \centering
    \includegraphics[width=\textwidth]{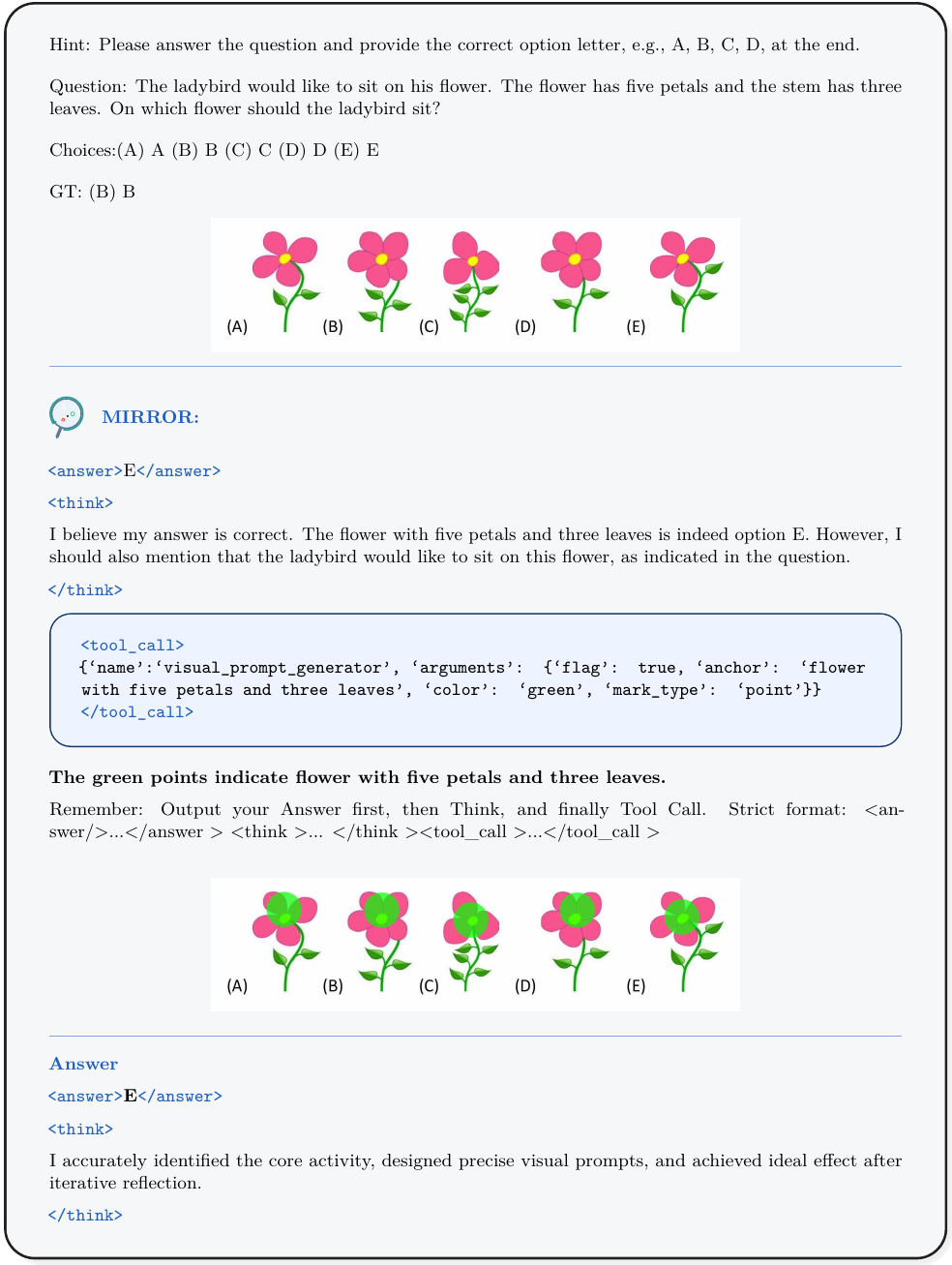} 
    \caption{\textbf{Coarse-grained Attribute Binding: }This case demonstrates the limitations of the current visual verification mechanism. 
    }
    \label{fig:failure_case2}
\end{figure*}

\end{document}